\documentclass{article}

\usepackage{microtype}
\usepackage{graphicx}
\usepackage{subcaption}
\usepackage{booktabs} 

\usepackage{hyperref}

\usepackage[preprint]{icml2026}

\usepackage{amsmath}
\usepackage{amssymb}
\usepackage{mathtools}
\usepackage{amsthm}
\usepackage{caption}
\usepackage{makecell}
\usepackage{array}
\usepackage{booktabs}
\usepackage{multirow}

\usepackage[capitalize,noabbrev]{cleveref}

\theoremstyle{plain}

\theoremstyle{definition}

\theoremstyle{remark}

\usepackage[textsize=tiny]{todonotes}

\icmltitlerunning{Generative Control as Optimization}

\begin{document}

\twocolumn[
  \icmltitle{Generative Control as Optimization: \\
  Time Unconditional Flow Matching for Adaptive and Robust Robotic Control}



  \icmlsetsymbol{equal}{*}

  \begin{icmlauthorlist}
    \icmlauthor{Zunzhe Zhang}{equal,thu,xht}
    \icmlauthor{Runhan Huang}{equal,thu}
    \icmlauthor{Yicheng Liu}{thu,xht}
    \icmlauthor{Shaoting Zhu}{thu}
    \icmlauthor{Linzhan Mou}{pton}
    \icmlauthor{Hang Zhao}{thu,xht}
  \end{icmlauthorlist}

  \icmlaffiliation{thu}{Institute for Interdisciplinary Information Sciences, Tsinghua University, Beijing, China}
  \icmlaffiliation{pton}{Department of Computer Science, Princeton University, New Jersey, US}
  \icmlaffiliation{xht}{Galaxea Inc., Beijing, China}

  \icmlcorrespondingauthor{Hang Zhao}{hangzhao@mail.tsinghua.edu.cn}

  \icmlkeywords{Machine Learning, ICML}

  \vskip 0.3in
]



\printAffiliationsAndNotice{* Equal contribution. Author order was determined randomly.}  

\begin{abstract}
Diffusion models and flow matching have become a cornerstone of robotic imitation learning, yet they suffer from a structural inefficiency where inference is often bound to a fixed integration schedule that is agnostic to state complexity. This paradigm forces the policy to expend the same computational budget on trivial motions as it does on complex tasks. We introduce \textbf{Generative Control as Optimization (GeCO)}, a time-unconditional framework that transforms action synthesis from trajectory integration into iterative optimization. GeCO learns a stationary velocity field in the action-sequence space where expert behaviors form stable attractors. Consequently, test-time inference becomes an adaptive process that allocates computation based on convergence—exiting early for simple states while refining longer for difficult ones. Furthermore, this stationary geometry yields an intrinsic, training-free safety signal, as the field norm at the optimized action serves as a robust out-of-distribution (OOD) detector, remaining low for in-distribution states while significantly increasing for anomalies. We validate GeCO on standard simulation benchmarks and demonstrate seamless scaling to $\pi_0$-series Vision-Language-Action (VLA) models. As a plug-and-play replacement for standard flow-matching heads, GeCO improves success rates and efficiency with an optimization-native mechanism for safe deployment. Video and code can be found at \href{https://hrh6666.github.io/GeCO/}{https://hrh6666.github.io/GeCO/}.
\end{abstract}

\vspace{-25pt}
\section{Introduction}
\label{sec:intro}

Generative modeling has rapidly established itself as a cornerstone of modern robotic imitation learning \cite{black2024pi_0, chi2025diffusion, janner2022planning, huang2025diffuse, huang2025flexible, ze20243d, kim2024openvla, bu2025univla}. By representing policies as conditional distributions rather than deterministic mappings, approaches such as diffusion models \cite{ho2020denoising, song2020denoising, zhang2023adding} and flow matching \cite{lipman2022flow, liu2022flow} have demonstrated exceptional capabilities in capturing the multi-modal nature of human behavior. These methods excel at synthesizing complex trajectories for manipulation and locomotion tasks \cite{chi2025diffusion, intelligence2025pi_, huang2025diffuse, ze20243d, ajay2022conditional}, driving state-of-the-art performance across diverse benchmarks \cite{chen2025robotwin, liu2023libero, zhang2025vlabench}. Their success stems largely from iteratively refining noise into feasible actions via time-conditioned dynamics and effectively decomposing the challenging generation problem into a sequence of manageable denoising or flow integration steps \cite{ho2020denoising, lipman2022flow}.

Yet, this reliance on time-conditioning introduces a fundamental structural inefficiency for control applications, a limitation that persists across standard formulations. These methods learn dynamic vector fields that evolve according to a pre-defined fictitious time schedule \cite{wang2025equilibrium}: while this schedule bridges Gaussian noise and the expert data distribution, it creates a rigid inference paradigm that is agnostic to the actual complexity of the current robot state. The policy is compelled to execute a fixed number of integration steps regardless of convergence, thereby wasting computational resources on trivial motions while potentially failing to sufficiently refine actions for complex tasks. This ``blind integration" problem not only hinders efficiency but also obscures the geometric structure of the policy. Time-varying fields lack a stationary energy landscape \cite{wang2025equilibrium, sun2025noise}: with the field direction shifting across timesteps, there is no intrinsic mechanism to verify action validity or detect out-of-distribution (OOD) scenarios, a critical gap for safe real-world deployment \cite{hodge2025out}.

To address these limitations, we reimagine generative control through a \textbf{time-unconditional} lens, grounding action synthesis in iterative optimization rather than trajectory integration. Our framework, \textbf{Generative Control as Optimization (GeCO)}, embodies a paradigm shift: instead of learning time-dependent dynamics, we learn a single, stationary velocity field in the action-sequence space where expert behaviors form stable attractors. To ensure robust convergence in continuous action spaces, we incorporate a velocity rescaling mechanism that modulates flow magnitude based on distance to the expert manifold, creating a geometric sink around target modes. This transforms inference from a fixed-schedule process into an adaptive optimization loop that naturally settles at equilibrium.

\begin{figure*}[t] 
    \centering
    
    \includegraphics[width=1.0\linewidth]{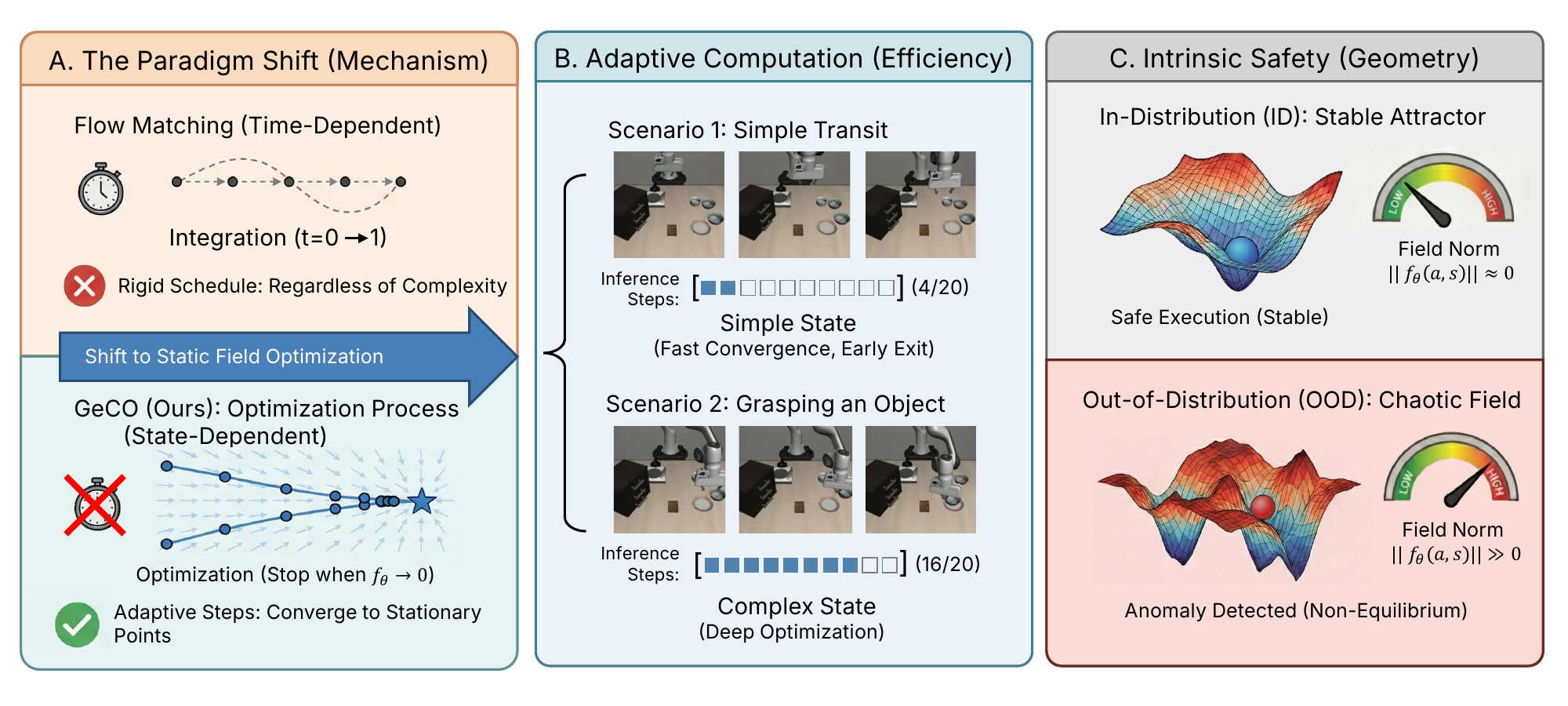}
    
    \caption{\textbf{Generative Control as Optimization (GeCO).} 
    \textbf{(A) The Paradigm Shift:} Unlike standard flow matching which relies on rigid, time-dependent integration schedules (top), GeCO learns a stationary velocity field where inference becomes an iterative optimization process toward stable attractors (bottom). 
    \textbf{(B) Adaptive Computation:} This formulation enables the policy to dynamically allocate computational budget based on state complexity—exiting early for simple transit phases (Scenario 1) while performing deep refinement for precise manipulation (Scenario 2). 
    \textbf{(C) Intrinsic Safety:} The stationary geometry provides a zero-shot safety mechanism. In-distribution (ID) states converge to low-energy equilibria ($||f_\theta|| \approx 0$), whereas out-of-distribution (OOD) anomalies exhibit persistently high field norms ($||f_\theta|| \gg 0$), enabling robust detection.}
    \label{fig:adaptive_profile}
    \vspace{-10pt}
\end{figure*}

The stationary nature of GeCO’s velocity field unlocks two transformative capabilities absent in time-dependent baselines, directly addressing our core motivations of adaptive computation and intrinsic OOD awareness. First, it enables \textbf{adaptive inference} rooted in convergence: rather than adhering to an arbitrary step count, the policy dynamically adjusts computation by exiting early for simple states (e.g., basic reaching) and spending more time refining actions for high-precision tasks (e.g., delicate manipulation). This decouples planning horizon from training schedules, allocating computational resources efficiently. Second, and crucially, the static geometric structure provides \textbf{intrinsic OOD awareness} as a training-free safety signal. The residual norm of the velocity field acts as a robust epistemic uncertainty metric: in-distribution states converge to zero-velocity attractors (low norm), while OOD states, lacking a learned manifold to settle on, exhibit persistent, non-vanishing gradients. This allows the robot to detect anomalies directly from the field’s energy, without auxiliary networks or ensembles.

Notably, GeCO exhibits strong architectural versatility, seamlessly supporting both standard Diffusion Transformers (DiT) \cite{chi2025diffusion, dong2024cleandiffuser, dong2025conditioning} and large-scale Vision-Language-Action (VLA) foundation models \cite{black2024pi_0, intelligence2025pi_}. A defining advantage lies in its \textbf{plug-and-play} design: it serves as a direct drop-in replacement for standard flow-matching heads \cite{black2024pi_0}, requiring minimal architectural adjustments or modifications to the loss function to integrate into existing systems. Building on this versatility, we conduct extensive validation of GeCO on standard simulation benchmarks \cite{chen2025robotwin, liu2023libero, zhang2025vlabench}. Empirically, GeCO outperforms state-of-the-art time-conditioned baselines while fundamentally redefining inference efficiency through the modulation of computational resources based on task complexity, rather than adherence to rigid fixed schedules. Beyond pure performance gains, we further establish that the intrinsic geometric signals of GeCO’s stationary field enable robust detection of task-level distribution shifts, delivering an optimization-native safety mechanism critical for reliable real-world robotic deployment.

\section{Related Work}
\textbf{Generative Models for Robotic Control.} Generative policy learning has become a dominant paradigm for robotic control in recent years. One prominent direction leverages the reasoning and perceptual capabilities of Vision-Language Models (VLMs) \cite{beyer2024paligemma, bai2023qwen, liu2023visual} for high-level planning \cite{driess2023palm, liang2022code, tian2024drivevlm} and action generation \cite{kim2024openvla,kim2025fine, bu2025univla}. Parallel efforts utilize video generation models \cite{liu2024sora, yang2024cogvideox, wiedemer2025video} to serve as predictive priors for robotic decision-making \cite{du2023learning, chen2025large, shen2025videovla}. In the realm of precise robotic control, diffusion \cite{ho2020denoising, song2020denoising} and flow matching \cite{lipman2022flow} have demonstrated exceptional effectiveness, functioning as robust planners and action-chunk generators \cite{chi2025diffusion, huang2025flexible, zhou2024diffusion, huang2025diffuse, janner2022planning, ajay2022conditional, dong2025conditioning, ze20243d}. Notably, the integration of continuous flow-matching heads into VLA architectures has emerged as a powerful paradigm \cite{black2024pi_0, intelligence2025pi_, li2025gr, shukor2025smolvla, jiang2025galaxea}; this approach bridges the semantic understanding of VLM backbones with high-frequency, continuous action execution. Such continuous generation mechanisms are widely adopted in modern Vision-Language-Action systems to replace purely autoregressive, discretized action decoding \cite{kim2024openvla, kim2025fine}, enabling higher fidelity control and improved real-time performance \cite{black2025real, intelligence2025pi_}. Our work focuses on this continuous generative control family of flow matching, which continues to drive advancements across domains ranging from robotic manipulation to navigation.

\textbf{Time-Unconditional Generative Dynamics.} Most diffusion and flow-based models \cite{ho2020denoising, song2020denoising, lipman2022flow} formulate generation as a \emph{time-conditioned} dynamical system, where the vector field varies with an explicit noise or time variable, typically necessitating a fixed integration horizon. Recent studies \cite{sun2025noise} indicate that removing this explicit conditioning can lead to graceful degradation, or even performance gains, suggesting that dynamics can often be inferred implicitly from the state. However, naively discarding time conditioning does not guarantee stable equilibrium behavior or well-posed optimization dynamics \cite{wang2025equilibrium}. \cite{wang2025equilibrium} address this by learning a stationary vector field compatible with an implicit energy landscape \cite{lecun2006tutorial}, enabling optimization-driven sampling \cite{florence2022implicit}. While this paradigm shows promise in image generation, its effectiveness in robotic control, which demands rigorous accuracy and sequential stability, still remains an open question.

\section{Method}
\label{sec:method}

Our goal is to turn generative control for robotic manipulation from a fixed-time integration process into an optimization problem over a time-invariant velocity field.
At each control step, the policy observes images, proprioception, and a language instruction, and synthesizes a short-horizon action sequence.
We first revisit time-conditioned diffusion and flow matching for such control, then introduce our time-unconditional field and optimization-based inference, and finally show how the same geometry yields intrinsic OOD awareness.

\subsection{Revisiting Time-Conditioned Generative Control}
\label{subsec:time-cond}

Diffusion- and flow-based controllers for manipulation typically learn a \emph{time-conditioned} velocity field
\[
v_\theta(x_\gamma, \gamma, s_t),
\]
where $x_\gamma$ is a noisy or partially denoised action sequence, $\gamma$ is a time variable, and $s_t$ is the current observation and instruction.
Inference follows an \emph{integration path in the time domain}: starting from noise, the policy traverses $\gamma$ from $0$ to $1$ using steps $\{\Delta\gamma_k\}_k$ with
\(
\sum_k \Delta\gamma_k \approx 1.
\)

Although one can change the number of steps and step sizes at deployment, these choices are constrained by the requirement of representing a valid integration of the time-conditioned field over $\gamma \in [0,1]$.
Stopping too early under-integrates the ODE/SDE; letting $\sum_k \Delta\gamma_k > 1$ pushes the system beyond the interval on which the field is defined and trained.
Thus the geometry of $v_\theta(\cdot,\gamma,s_t)$ is always tied to a specific notion of time progress.

This design couples three roles into a single scalar~$\gamma$: tracking progress along the generative trajectory, controlling the strength of the field, and parameterizing the integration domain.
As a result, the field geometry changes with~$\gamma$, and there is no single, time-invariant notion of ``good actions'' or a creteria whose magnitude directly measures how close a final action is to the in-distribution manifold.

GeCO decouples this dependence on time: instead of conditioning on $\gamma$, we learn a \emph{stationary} field over action sequences and use it as the objective of an unconstrained optimization procedure.

\subsection{Time-Unconditional Velocity Fields for Control}
\label{subsec:uncond-field}

At timestep $t$, the robot observes $s_t$ (multi-view RGB of the workspace, proprioception, and a language instruction) and predicts a sequence of future low-level actions.
\[
a = a_{t:t+T_a-1} \in \mathbb{R}^{T_a \times d_a},
\]

Rather than a time-indexed $v_\theta(\cdot,\gamma,s_t)$, GeCO learns a \emph{time-unconditional} velocity field $f_\theta(x, s_t)$ over action sequences. We want $f_\theta(\cdot,s_t)$ to point from noisy or suboptimal sequences toward \emph{equilibrium actions} that solve the current task, and to vanish near ground truth in-distribution actions, in order stop automatically during inference without an explicit time condition.

To train such a field, we introduce $\gamma$ only as a implicit training variable.
Given a demonstration $a$ and Gaussian noise $\varepsilon \sim \mathcal{N}(0,I)$, we form
\begin{equation}
x_\gamma = \gamma a + (1 - \gamma)\varepsilon,
\qquad
\gamma \sim \mathcal{U}(0,1),
\end{equation}
and crucially \emph{do not} provide $\gamma$ to the model.
At each $x_\gamma$ we define a restoring direction
\begin{equation}
g^\star(a,\varepsilon,\gamma)
=
(\varepsilon - a)\,c(\gamma),
\label{eq:target-field}
\end{equation}
where $c(\gamma)$ is a scalar schedule that decays to zero as $\gamma \to 1$.
Because $c(1)=0$, the target field vanishes at $x_\gamma = a$, transforming ground-truth action sequences into natural stationary equilibrium points of the learned field. Given such rescaling mechanism, the denoising process is able to automatically stop in the ground truth stationary points, effectively solving the non-equilibrium nature of the vanilla time-unconditional flow matching. 

Conditioned on $s_t$, the loss is
\begin{equation}
\mathcal{L}(\theta)
=
\mathbb{E}_{(s_t, a)}
\mathbb{E}_{\varepsilon, \gamma}
\Bigl\|
f_\theta(x_\gamma, s_t) - g^\star(a,\varepsilon,\gamma)
\Bigr\|^2.
\label{eq:geco-loss}
\end{equation}
Here $\gamma$ only controls the interpolation and the scale via $c(\gamma)$; since it never enters the model, $f_\theta(\cdot,s_t)$ must organize all training pairs into a \emph{single} consistent field for each scene.
This encourages $f_\theta$ to approximate the gradient of an implicit potential field over action sequences whose minima lie near successful demonstrations.

\subsection{Optimization-Based Inference}
\label{subsec:opt-inference}

Once $f_\theta$ has been learned, inference is no longer framed as integrating over a time axis.
At deployment, the controller runs in closed loop: at each timestep $t$, it observes $s_t$ and solves a small optimization problem in action space using $f_\theta(\cdot,s_t)$.

We initialize $a^{(0)}$ from a prior $p_{\text{prior}}(\cdot \mid s_t)$, e.g., isotropic Gaussian noise, and then perform $K$ gradient-based updates:
\begin{equation}
a^{(k+1)}
=
a^{(k)} - \eta_k\, f_\theta(a^{(k)}, s_t),
\label{eq:gd-update}
\end{equation}
where $\eta_k$ may depend on the sample or iteration.
Because there is no time index, and there is no requirement that the effective ``total step length" $\sum_k \eta_k$ equal a particular value.
Taking more or fewer steps, or even $\sum_k \eta_k > 1$, remains a natural optimization process in the same stationary field, rather than violating a time-integration constraint and leading to non-equilibrium divergence.

Convergence is monitored directly through the field:
\begin{equation}
\|f_\theta(a^{(k)}, s_t)\| \le \tau_{\text{opt}}
\quad \text{or} \quad
k \ge K_{\max}.
\label{eq:stopping}
\end{equation}
Simple scenes (e.g., unobstructed reaches) reach a low-norm region in few iterations; cluttered or contact-rich scenes require more.
GeCO behaves as an optimization-based, receding-horizon manipulation controller grounded in a learned generative field. Detailed inference pseudocode can be found in Algo.\ref{algo:infer}.

\begin{algorithm}[tb]
\caption{Adaptive Inference (GeCO)}
\label{algo:infer}
\begin{algorithmic}[1]
\STATE {\bfseries Input:} Observation $s$, Max steps $K$, Tolerance $\tau$, Rate $\eta$
\STATE {\bfseries Output:} Action $\hat{a}$, OOD Score $S$
\STATE $a^{(0)} \sim \mathcal{N}(0, I)$ \hfill $\triangleright$ \textit{Initialize prior}
\FOR{$k = 0, \dots, K-1$}
    \STATE $v_k \leftarrow f_\theta(a^{(k)}, s)$
    \IF{$\| v_k \|_2 < \tau$}
        \STATE \textbf{break} \hfill $\triangleright$ \textit{Adaptive early exit}
    \ENDIF
    \STATE $a^{(k+1)} \leftarrow a^{(k)} - \eta \cdot v_k$
\ENDFOR
\STATE $\hat{a} \leftarrow a^{(k)}$
\STATE \textbf{return} $\hat{a}$, $\|f_\theta(\hat{a}, s)\|_2$
\end{algorithmic}
\end{algorithm}

\subsection{Intrinsic OOD Detection from Field Geometry}
\label{subsec:ood-detector}

GeCO also provides \emph{optimization-native} OOD awareness.
Here OOD is defined at the level of the conditional field:
training exposes the model to $(s_t,a)$ pairs from $p_{\text{data}}(s_t,a)$, and thus to conditional fields $f_\theta(\cdot,s_t)$ induced by $s_t \sim p_{\text{data}}(s_t)$.
At test time, an observation $s_t$ is OOD if it induces a field whose geometry is inconsistent with this family; optimization on such a field need not reach a low-norm equilibrium, and this breakdown in convergence becomes an OOD signal.


Given a test observation $s_t$, we run optimization as in~\eqref{eq:gd-update}–\eqref{eq:stopping}, obtaining an approximate equilibrium $\hat{a}(s_t)$, and define
\begin{equation}
\mathrm{Score}(s_t)
=
\bigl\|f_\theta(\hat{a}(s_t),\, s_t)\bigr\|.
\label{eq:ood-score}
\end{equation}

For in-distribution robot control states $s_t \sim p_{\text{data}}$, the induced field resembles those seen during training: optimization can move $x_\gamma$ into regions that behave like high-$\gamma$ samples, where $c(\gamma)$ is small and the field is trained to nearly vanish, so $\mathrm{Score}(s_t)$ is low.
When $s_t$ is OOD, the induced field is itself out-of-distribution: optimization operates in regions never trained to lead to small norm, the gradient norm at the end of the optimization process at $\hat{a}(s_t)$ remains large.
We obtain a binary OOD decision via
\begin{equation}
\mathrm{isOOD}(s_t)
=
\mathbf{1}\{\mathrm{Score}(s_t) > \tau_{\text{OOD}}\}.
\label{eq:ood-threshold}
\end{equation}
This incurs essentially no overhead, as the same gradient norms are already computed for convergence checking.
\begin{figure*}[t]
    \centering
    
    \begin{subfigure}[b]{0.16\linewidth}
        \centering
        \includegraphics[width=\linewidth]{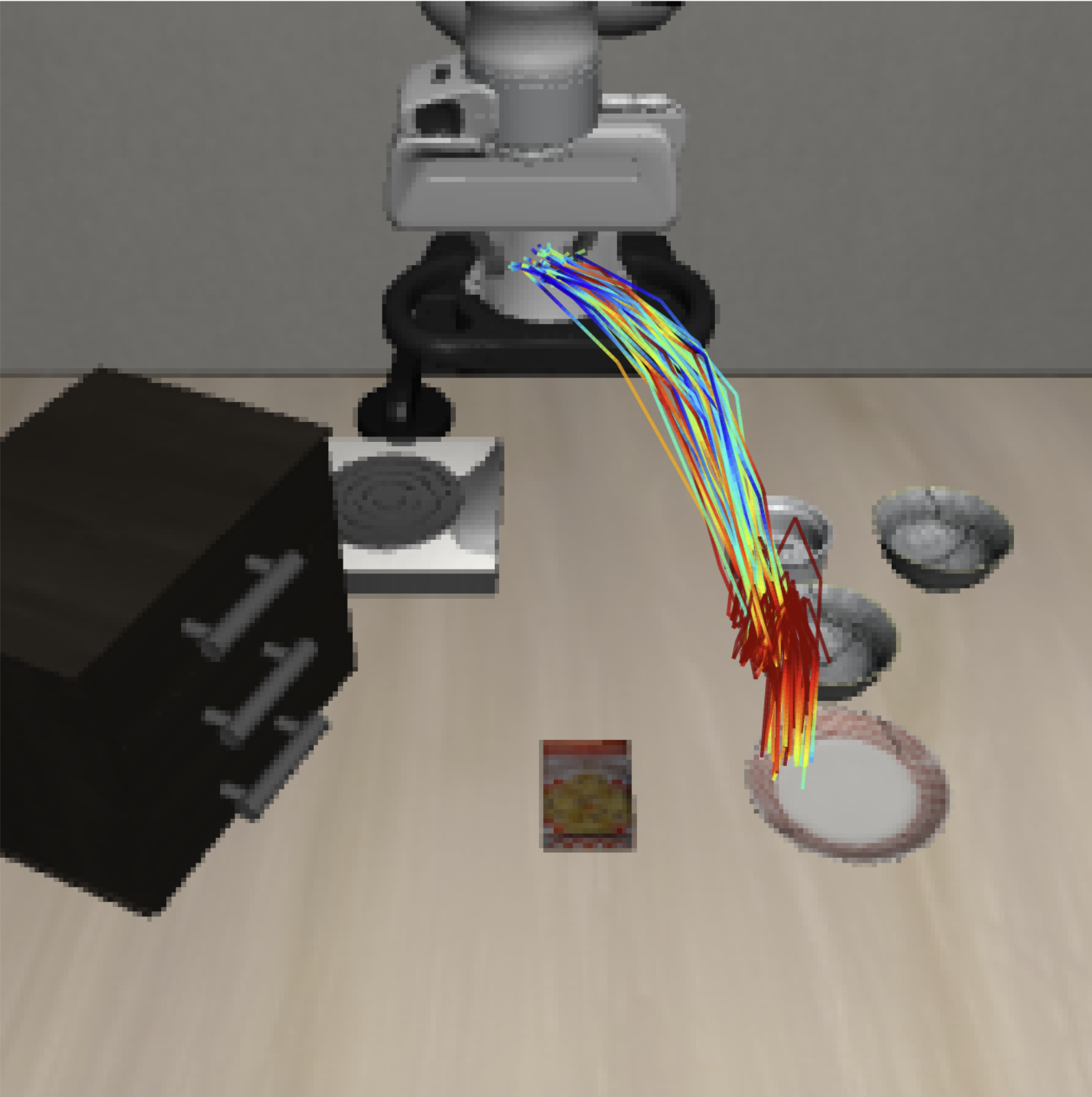}
        \caption{} 
        \label{fig:sp0}
    \end{subfigure}
    \hfill 
    \begin{subfigure}[b]{0.16\linewidth}
        \centering
        \includegraphics[width=\linewidth]{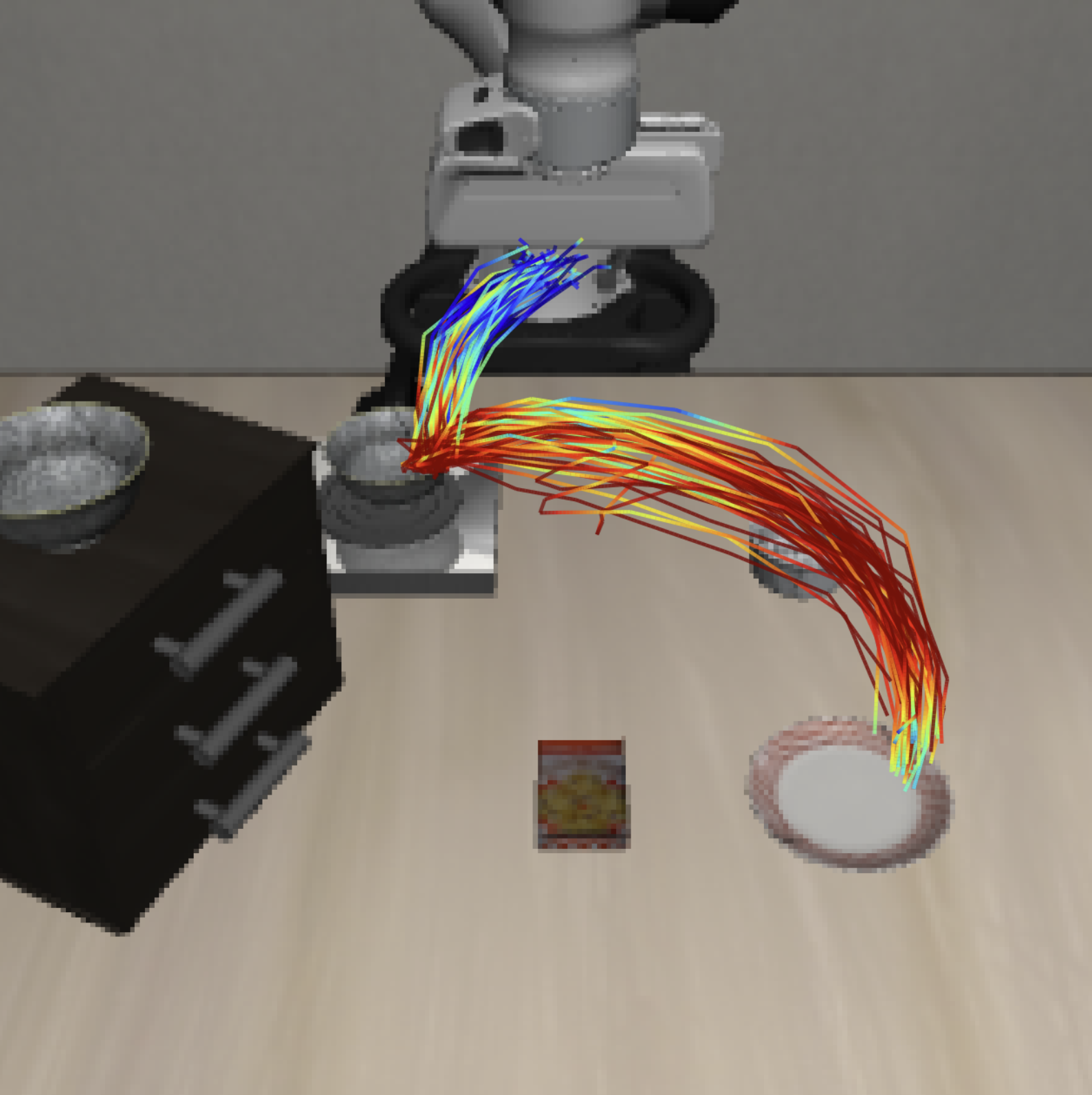}
        \caption{}
        \label{fig:sp1}
    \end{subfigure}
    \hfill
    \begin{subfigure}[b]{0.16\linewidth}
        \centering
        \includegraphics[width=\linewidth]{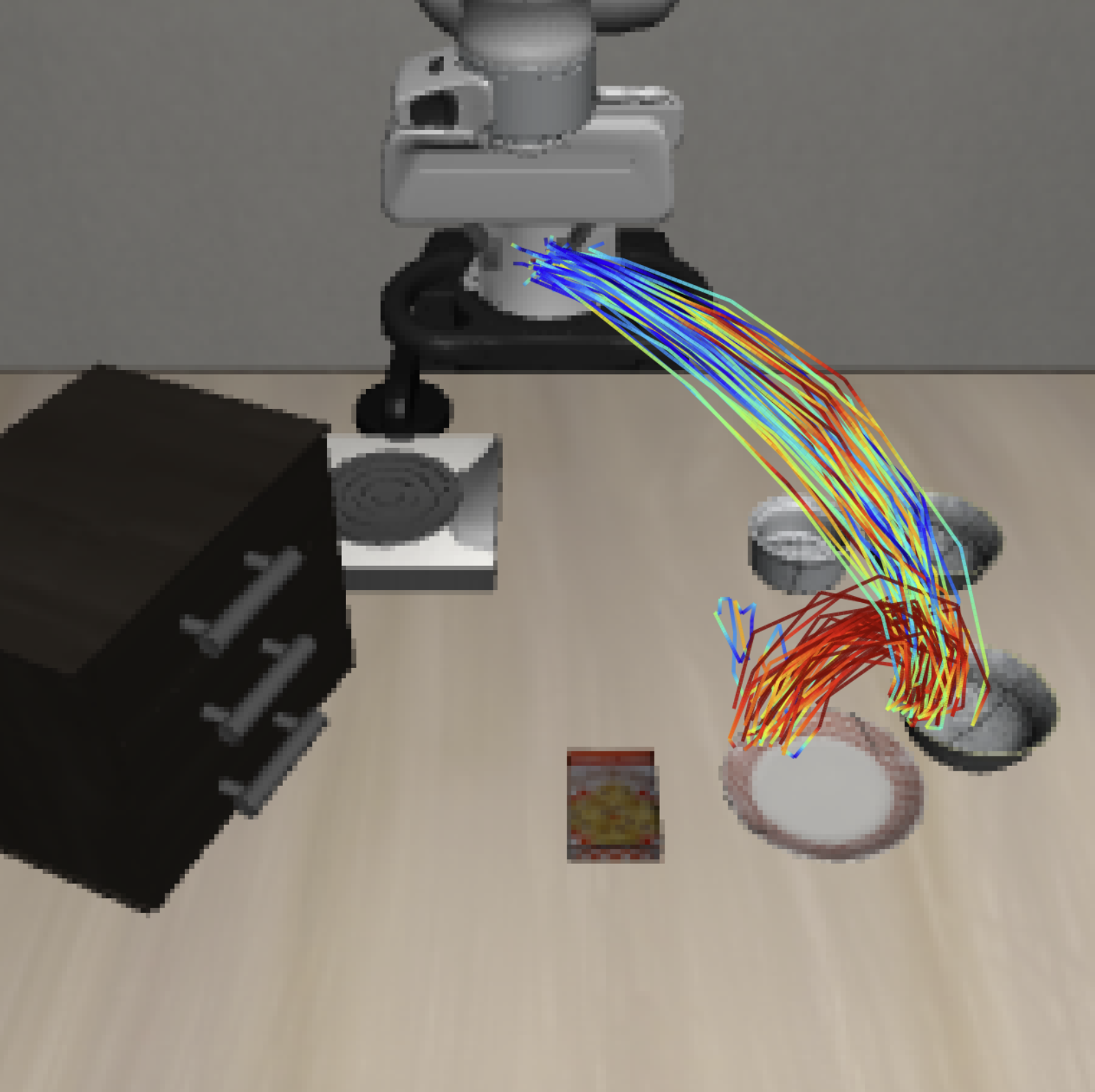}
        \caption{}
        \label{fig:sp2}
    \end{subfigure}
    \hfill
    \begin{subfigure}[b]{0.16\linewidth}
        \centering
        \includegraphics[width=\linewidth]{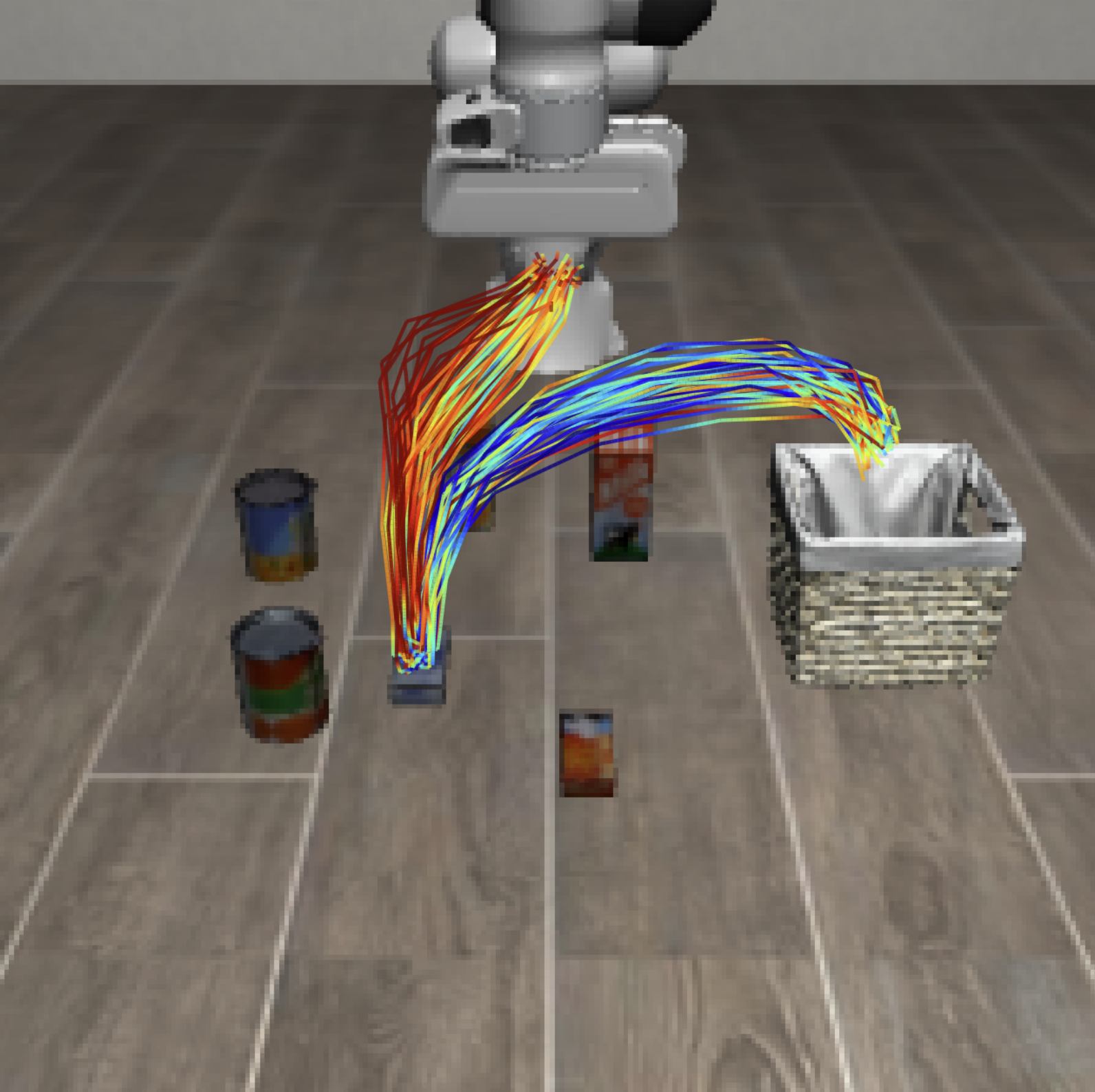}
        \caption{}
        \label{fig:ob1}
    \end{subfigure}
    \hfill
    \begin{subfigure}[b]{0.16\linewidth}
        \centering
        \includegraphics[width=\linewidth]{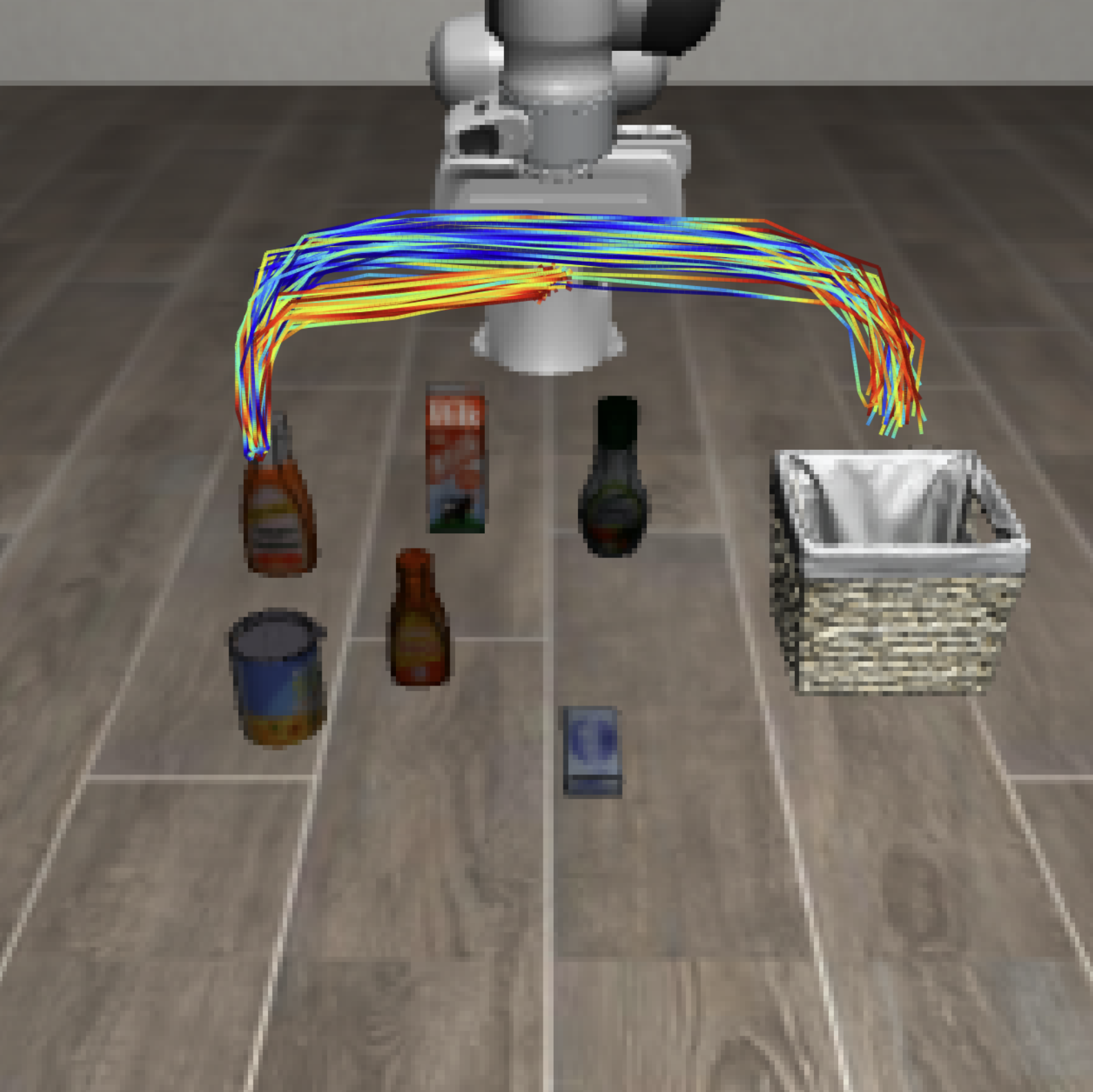}
        \caption{}
        \label{fig:ob4}
    \end{subfigure}
    \hfill
    \begin{subfigure}[b]{0.16\linewidth}
        \centering
        \includegraphics[width=\linewidth]{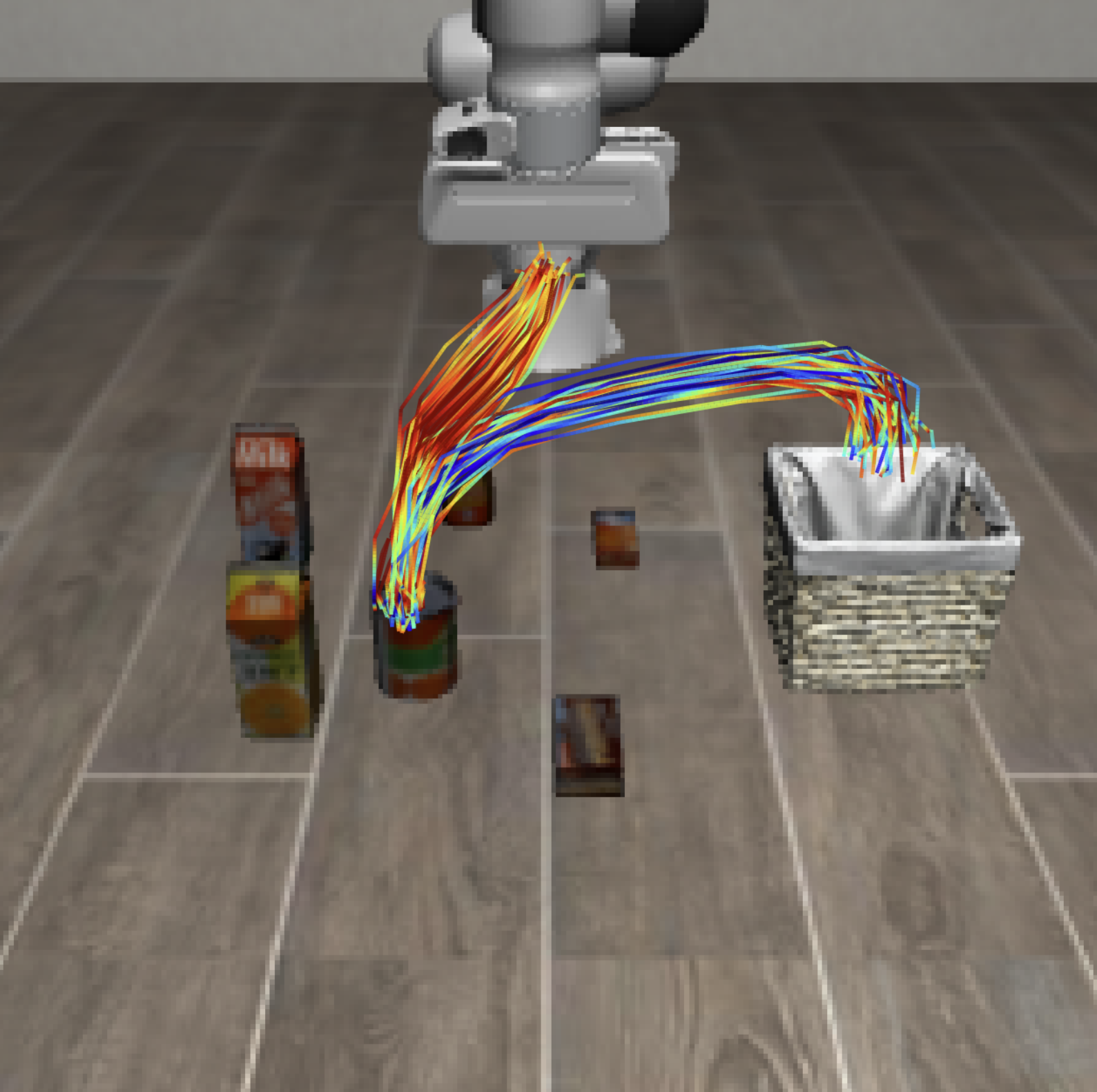}
        \caption{}
        \label{fig:ob5}
    \end{subfigure}

    \caption{Computation Follows Task Complexity.
We visualize the spatial distribution of inference effort along a single rollout. The first three panels (a–c) are sampled from LIBERO-Spatial, and the last three panels (d–f) are from LIBERO-Object. The color of each line encodes the number of function evaluations (NFE) required for convergence at that state, ranging from blue (NFE = 1) to red (NFE = 20). This visualization illustrates how GeCO allocates more computation to challenging states while using fewer steps in easier regions.}
    \label{fig:adaptive_visualization}
\end{figure*}
\subsection{Plug-and-Play Integration with VLA Systems}
\label{subsec:vla-integration}

Modern flow-based Vision-Language-Action (VLA) systems \cite{black2024pi_0, intelligence2025pi_, jiang2025galaxea, shukor2025smolvla} consist of two core components: (1) a Vision-Language Model (VLM) that fuses multi-modal inputs (multi-view RGB $\mathbf{v} \in \mathbb{R}^{H \times W \times 3 \times N_v}$, proprioception $\mathbf{p} \in \mathbb{R}^{d_p}$, language instruction $l \in \mathcal{L}$) into a task-aware conditional signal $s_t = \text{VLM}(\mathbf{v}, \mathbf{p}, l)$; (2) a time-conditioned flow-matching head that takes $s_t$ and action sequence $x_\gamma$ as inputs to predict $v_\theta(x_\gamma, \gamma, s_t)$.

GeCO enables plug-and-play integration with such VLA architectures via minimal modifications: we directly replace the time-conditioned flow-matching head $v_\theta(x_\gamma, \gamma, s_t)$ with our time-unconditional velocity field $f_\theta(x, s_t)$ (Section \ref{subsec:uncond-field}), while retaining the VLM's multi-modal fusion pipeline and the VLA system's original input-output interface. No architectural changes to the VLM or control pipeline are required—at inference, the VLA system generates action sequences $\mathbf{a} \in \mathbb{R}^{T_a \times d_a}$ via GeCO's adaptive optimization loop (Algorithm \ref{algo:infer}) using the same $s_t$ signal, leveraging the stationary geometry of $f_\theta$ to eliminate the rigid time-schedule constraints of baseline flow-matching heads.

\section{Experiments}
\label{sec:experiments}

We evaluate \textbf{GeCO} along three complementary axes: 
(1) \textbf{Adaptive Efficiency}, analyzing how the optimization-based inference dynamically allocates computation based on task complexity;
(2) \textbf{Scalability}, assessing the method's effectiveness as a plug-and-play head for large Vision-Language-Action (VLA) models ($\pi_0$ series); and 
(3) \textbf{Intrinsic Safety}, verifying the reliability of the stationary field norm as a zero-shot Out-of-Distribution (OOD) detector. Implementation details can be found in Appendix \ref{sec:appendix_impl_adaptive}.

\subsection{Policy Performance and Adaptive Computation}
\label{subsec:adaptive_comp}

\paragraph{The Efficiency-Performance Trade-off.}
Table~\ref{tab:libero_main} demonstrates that GeCO achieves a superior trade-off between inference speed and task success compared to fixed-schedule baselines on the LIBERO \cite{liu2023libero} benchmark. Remarkably, GeCO with a budget of only 5 steps already outperforms Rectified Flow (20 steps), achieving a higher success rate (91.9\% vs. 90.0\%) while requiring \textbf{75\% less compute}. When the budget is increased to 20 steps, GeCO reaches a peak success rate of 93.5\%. Crucially, the average NFE required to achieve this peak is only 11.6, significantly lower than the maximum budget. This indicates that GeCO's optimizer effectively allocates computational resources, autonomously exiting early for simpler states while reserving the budget for complex ones.

\paragraph{Temporal Analysis: Computation Follows Complexity.}
To verify that the observed variation in NFE is structurally driven by task complexity rather than randomness, we analyze the inference profile shown in Figure~\ref{fig:adaptive_visualization}. We observe distinct optimization behaviors across different task types:
\begin{itemize}
    \item \textbf{Motion Planning Complexity (LIBERO-Spatial):} For tasks requiring spatial reorientation (Fig.~\ref{fig:adaptive_visualization} a-c), higher NFE concentrates heavily around bottleneck phases, such as pre-grasp alignment and object placement. In contrast, free-space transit phases converge rapidly, demonstrating that the learned vector field provides a smooth gradient in unobstructed regions.
    \item \textbf{Semantic Grounding (LIBERO-Object):} 
In tasks where the robot must identify a specific target among multiple distractors based on language instructions (e.g. pick the ketchup and place it in the basket), we observe a distinct NFE spike at the episode onset (Fig.~\ref{fig:adaptive_visualization} d-f). 
This suggests that the optimizer requires more iterations to resolve the visual-semantic correspondence, effectively deliberating to lock onto the specific object defined by the instruction before committing to a trajectory.
\end{itemize}
These results confirm that GeCO acts as an adaptive controller, dynamically modulating its inference horizon based on both kinematic constraints and perceptual uncertainty.
\begin{table}[H]
\centering
\caption{\textbf{Performance vs. Efficiency on LIBERO.} GeCO achieves higher success rates by allowing adaptive refinement. By increasing the max step budget, GeCO refines actions for complex states, yet the average compute remains low compared to fixed-schedule baselines.}
\label{tab:libero_main}
\vskip 0.1in
\begin{small}
\resizebox{\columnwidth}{!}{%
    \setlength{\tabcolsep}{2.5pt} 
    \begin{tabular}{l|c|ccccc|c}
    \toprule
    \multirow{2}{*}{Method} & Max & \multicolumn{5}{c|}{Success Rate (\%)} & \multicolumn{1}{c}{Efficiency} \\
    & Steps & Goal & Spat. & Obj. & Long & Avg. & \textbf{NFE} \\
    \midrule
    Diff. Policy & 100 (Fixed) & 82.9 & 91.4 & 88.9 & 82.7 & 86.5 & 100.0 \\
    Rectified flow & 20 (Fixed) & 92.4 & 94.6 & 97.0 & 76.0 & 90.0 & 20.0 \\
    \midrule
    \textbf{GeCO (Ours)} & 5 & 91.6 & 95.4 & 98.2 & 82.4 & 91.9 & \textbf{5.0} \\
    \textbf{GeCO (Ours)} & 10 & 93.0 & 95.8 & 99.0 & 81.8 & 92.4 & 8.7 \\
    \textbf{GeCO (Ours)} & 20 & \textbf{95.2} & \textbf{95.8} & \textbf{99.0} & 83.8 & \textbf{93.5} & 11.6 \\
    \textbf{GeCO (Ours)} & 30 & 93.6 & 95.2 & 98.8 & \textbf{84.8} & 93.1 & 12.8 \\
    \bottomrule
    \end{tabular}%
}
\end{small}
\end{table}

\begin{table*}[t]
\centering
\caption{
Success rate (\%) of different VLA methods on the 
\textbf{RoboTwin 2.0} benchmark. Each task is evaluated under two difficulty 
levels: \textbf{Easy} and \textbf{Hard}. For every method we report the success 
rate for each task and difficulty level, together with the average performance 
across all five tasks. Higher values indicate better task completion ability. 
The best result in each column is highlighted in \textbf{bold}.
}
\label{tab:robotwin_benchmark_vla_eh}
\vspace{-2pt}

\begin{small}

\setlength{\tabcolsep}{1.4pt}
\renewcommand{\arraystretch}{1.08}

\newcolumntype{C}{>{\centering\arraybackslash}p{0.78cm}}

\begin{tabular}{l|CC CC CC CC CC|CC}
\toprule
\multirow{2}{*}{Method}
& \multicolumn{2}{c}{\makecell[c]{Adjust\\Bottle}}
& \multicolumn{2}{c}{\makecell[c]{Beat Block\\Hammer}}
& \multicolumn{2}{c}{\makecell[c]{Blocks Ranking\\(Size)}}
& \multicolumn{2}{c}{\makecell[c]{Click\\Alarmclock}}
& \multicolumn{2}{c}{\makecell[c]{Click\\Bell}}
& \multicolumn{2}{c}{Avg.} \\

\cmidrule(lr){2-3}
\cmidrule(lr){4-5}
\cmidrule(lr){6-7}
\cmidrule(lr){8-9}
\cmidrule(lr){10-11}
\cmidrule(lr){12-13}

& Easy & Hard
& Easy & Hard
& Easy & Hard
& Easy & Hard
& Easy & Hard
& Easy & Hard \\

\midrule

ACT \citep{zhao2023learning}
& \textbf{97.0} & 23.0
& \textbf{56.0} & 3.0
& 0.0 & 0.0
& 32.0 & 4.0
& \textbf{58.0} & 3.0
& 49.0 & 7.0 \\

$\pi_0$ \citep{black2024pi_0}
& 90.0 & 56.0
& 43.0 & 21.0
& \textbf{7.0} & \textbf{1.0}
& 63.0 & 11.0
& 44.0 & 3.0
& 49.0 & 18.0 \\

$\pi_0$ \textbf{with GeCO}
& 86.0 & \textbf{66.0}
& 52.0 & \textbf{36.0}
& 0.0 & 0.0
& \textbf{72.0} & \textbf{16.0}
& 42.0 & \textbf{21.0}
& \textbf{50.0} & \textbf{28.0} \\

\bottomrule
\end{tabular}

\end{small}

\vskip -0.08in

\end{table*}

\begin{table}[t]
\centering
\caption{VLA success rates (\%) on VLABench tracks. We report track-wise success and the average over the listed tracks; higher is better.}
\label{tab:vlabench_benchmark_vla}
\vspace{-2pt}
\begin{small}
\setlength{\tabcolsep}{2.4pt}
\renewcommand{\arraystretch}{1.08}

\begin{tabular}{l|ccccc|c}
\toprule
Method
& \makecell[c]{Track 1}
& \makecell[c]{Track 2}
& \makecell[c]{Track 3}
& \makecell[c]{Track 4}
& \makecell[c]{Track 6}
& Avg. \\
\midrule
$\pi_0$ & 47.0 & 21.2 & 29.1 & 17.3 & 32.2 & 29.4 \\
$\pi_{0.5}$ & 40.6 & \textbf{22.6} & 18.0 & 16.1 & 25.6 & 24.5 \\
\textbf{$\pi_0$ with GeCO} & \textbf{61.0} & 22.0 & \textbf{34.0} & \textbf{20.0} & \textbf{45.0} & \textbf{36.0} \\
\bottomrule
\end{tabular}

\end{small}
\vskip -0.08in
\end{table}

\subsection{Scalability to Vision-Language-Action Models}
\label{subsec:vla_scale}

We evaluate the scalability of GeCO by integrating it into flow-matching-based VLA models, with the $\pi_0$-series models selected as a representative example. A critical advantage of GeCO is its architectural compatibility with existing time-conditional backbones. Since $\pi_0$ is originally trained as a time-dependent flow matching model expecting a time embedding $t \in [0, 1]$, we adapt it to our time-unconditional framework by simply fixing the input time $t=0$ during fine-tuning. 
This effectively freezes the time channel, forcing the network to learn a single, stationary velocity field $f_\theta(x, s)$ using the pre-trained weights, rather than a sequence of time-varying fields. This plug-and-play adaptation integrate the rich visual-semantic representations of foundation models with the native adaptive optimization inference of GeCO, with no need for structural modifications or training from scratch. For a fair comparison, both GeCO and the baselines are fine-tuned from the same pre-trained $\pi_0$ base for 30,000 steps. Details can be found in Appendix \ref{sec:appendix_benchmarks}.
\paragraph{VLABench Result}
We benchmark our method on VLABench~\cite{zhang2025vlabench}, a challenging suite designed to evaluate long-horizon, language-conditioned robotic manipulation with strong requirements on semantic grounding and multi-step reasoning. As shown in Table~\ref{tab:vlabench_benchmark_vla}, $\pi_0$ with GeCO achieves the strongest overall performance with an average success rate of \textbf{0.36}, improving upon the baseline $\pi_0$ (0.294). The most substantial gains are observed on Track~1 and Track~6, which respectively evaluate in-distribution manipulation skills and robustness to unseen textures and visual variations. In addition, GeCO consistently improves results on Track~3 (common-sense and world knowledge) and Track~4 (semantic instruction following), both of which emphasize multi-step reasoning and long-horizon task execution. By contrast, performance on Track~2 (cross-category generalization) remains comparable to standard $\pi_0$, suggesting that improvements are more pronounced on execution- and reasoning-intensive tasks than on pure category-level transfer.
\paragraph{RoboTwin 2.0 Benchmark Result}
RoboTwin 2.0 \cite{chen2025robotwin} evaluates bimanual manipulation under two difficulty settings: \emph{Easy} uses clean environments, while \emph{Hard} tests the same policies under domain-randomized evaluation with clutter, texture, lighting, and tabletop-height variations.
As shown in Table~\ref{tab:robotwin_benchmark_vla_eh}, adding GeCO to $\pi_0$ substantially improves robustness in the Hard setting, raising the average success rate from 0.18 to \textbf{0.28}, while slightly improving the Easy average (0.49$\rightarrow$0.50).
The gains are consistent across several contact-rich tasks, including Adjust Bottle (0.56$\rightarrow$0.66), Beat Block Hammer (0.21$\rightarrow$0.36), Click Alarmclock (0.11$\rightarrow$0.16), and Click Bell (0.03$\rightarrow$0.21), indicating improved tolerance to visual and scene perturbations.
By contrast, Blocks Ranking (Size) remains challenging for all methods (near-zero success in both settings), suggesting that fine-grained ordering under bimanual coordination is still a major bottleneck on RoboTwin 2.0.
\paragraph{LIBERO Benchmark Result}
LIBERO \cite{liu2023libero} comprises four suites that isolate different distribution shifts in language-conditioned manipulation: goal changes (Goal), spatial relation shifts (Spat.), object category shifts (Obj.), and long-horizon compositional execution (Long).
In Table~\ref{tab:vla_scale}, GeCO remains comparable overall relative to standard fine-tuning: for $\pi_0$ it matches a similar average (93.9 vs. 94.2), and for $\pi_{0.5}$ it stays close (95.9 vs. 96.9).
These results suggest that GeCO maintains performance comparable to the flow matching baseline across different VLA model backbones.

Collectively, these results validate that GeCO acts as a better drop-in replacement for VLA action heads, delivering higher performance and robustness under identical training conditions.


\begin{table}[t]
\centering
\caption{VLA success rates (\%) on LIBERO benchmark. We report track-wise success and the average over the listed tracks; higher is better.}
\label{tab:vla_scale}
\vskip 0.15in
\begin{small}
\addtolength{\tabcolsep}{-3.5pt}
\begin{tabular}{l|cccc|c}
\toprule
Method & Goal & Spat. & Obj. & Long & Avg. \\
\midrule
$\pi_0$ + FAST \cite{pertsch2025fast} & 88.6 & 96.4 & 96.8 & 60.2 & 85.5 \\
$\pi_0$ \cite{black2024pi_0} & 95.8 & 96.8 & 98.8 & 85.2 & 94.2 \\
$\pi_{0.5}$ \cite{intelligence2025pi_} & \textbf{98.0} & \textbf{98.8} & 98.2 & \textbf{92.4} & \textbf{96.9} \\
\midrule
\textbf{Ours ($\pi_0$)} & 96.4 & 96.8 & 97.0 & \textbf{85.4} & 93.9 \\
\textbf{Ours ($\pi_{0.5}$)} & 96.4 & 97.4 & 98.6 & 92.4 & 96.2\\
\bottomrule
\end{tabular}
\end{small}
\vskip -0.1in
\end{table}
\subsection{Intrinsic OOD Detection from Optimization}
\label{subsec:ood_exp}

GeCO performs action synthesis via iterative optimization at each planning call. We hypothesize that under distribution shift, this optimization fails to converge cleanly, leading to larger update magnitudes. We therefore use the \textbf{final update norm} produced by a single planning call as an intrinsic anomaly score.

\paragraph{Experiment Setup.}
We train a single policy on \textit{LIBERO-Goal} (10 tasks) and evaluate OOD detection on \textit{LIBERO-Spatial} (10 tasks) as \textbf{Task OOD}. We choose LIBERO-Goal for training because tasks share the same scene layout and objects but differ in goal predicates, making performance sensitive to correctly grounding language instructions into behaviors. At test time, LIBERO-Spatial introduces unseen spatial arrangements of otherwise similar objects, creating a controlled workspace-layout shift while keeping the manipulation vocabulary comparable.

\paragraph{Baseline OOD Signal.}
Following the Diff-DAgger-style idea of using the diffusion training objective as an uncertainty signal~\cite{lee2025diff}, we define a lightweight baseline based on the \emph{flow-matching training loss}. The original formulation estimates an expectation over many sampled timesteps (e.g., 512 samples); to avoid extra sampling cost, we use a single-step proxy and evaluate the flow-matching loss at a fixed timestep $t=1$ for each planning call. This produces one scalar baseline score per planning call. 

\paragraph{Anomaly Score and AUROC Computation.}
An episode consists of multiple \emph{planning calls} (each call runs optimization to produce an action chunk that is then executed in the environment). We compute an anomaly score \emph{per planning call}. For the $j$-th planning call at state $s^{(j)}$, we run GeCO optimization for at most $K_{\max}=10$ steps. With the update
$a_{k+1} \leftarrow a_k - \eta\, f_\theta(a_k, s^{(j)})$,
we define
\[
\text{score}\bigl(\text{plan } j\bigr) \;=\; \bigl\| f_\theta(a^{(j)}_{K}, s^{(j)}) \bigr\|_2,
\]
where $K \le K_{\max}$ denotes the last executed refinement step (if early termination is triggered, $K$ is the stopping step; otherwise $K=K_{\max}$).
We sample 500 in-distribution (ID) episodes from LIBERO-Goal and 500 OOD episodes from LIBERO-Spatial. To avoid overweighting longer episodes, we compute AUROC on a fixed number of planning calls per episode, treating OOD plans as positives and ID plans as negatives.

\paragraph{Temporal Filtering and Early Reporting.}
Within each planning call, raw per-step norms can be noisy, so we evaluate two lightweight filters over the refinement-step residual norms
$r_k=\|f_\theta(a_k,s^{(j)})\|_2$.
Both filters output an \emph{online} OOD flag for the current planning call; once the flag triggers (at any planning call), we \textbf{immediately terminate the episode} to avoid executing the remaining (eventually failing) interactions.

\begin{itemize}
    \item \textbf{Moving Average.}
    We smooth $r_k$ by averaging over the last $w=5$ refinement steps,
    $
    \tilde{r}_k = \frac{1}{w}\sum_{i=k-w+1}^{k} r_i,
    $
    and trigger an OOD flag when $\tilde{r}_k$ exceeds a threshold.

    \item \textbf{Leaky Bucket.}
    We maintain an accumulator that integrates only sustained large residuals:
    $
    b_k \leftarrow \max\{0,\; (1-\lambda)b_{k-1} + (r_k-\tau)_+\},
    $
    where $(x)_+=\max(x,0)$.
    We raise an OOD flag if $b_k \ge B$ at any step.
\end{itemize}

\paragraph{Quantitative Results.}
Table~\ref{tab:ood_comparison} reports separability and operating-point performance for distinguishing ID (LIBERO-Goal) from OOD (LIBERO-Spatial).
We treat OOD plans as positives and ID plans as negatives; thus \textbf{TNR} measures the fraction of OOD instances correctly flagged, while \textbf{TNR} measures the fraction of ID instances correctly retained (i.e., not falsely flagged).
The baseline loss signal is less aligned with the suite-level shift, whereas GeCO yields strong separation.
We choose the detection threshold to target \textbf{TPR $\approx 90\%$} for our detector when possible, and report the corresponding TNR.

Time Saved measures the fraction of interaction time avoided on OOD episodes by early reporting: for each OOD episode, we record the first reporting time $t_{\mathrm{report}}$ (the environment time step when the first planning call triggers the OOD flag) and normalize by the full episode duration $T_{\mathrm{total}}$ under the standard evaluation protocol,
$\text{Time Saved}=1-t_{\mathrm{report}}/T_{\mathrm{total}}$; we then average this quantity over OOD episodes.

\begin{table}[t]
\centering
\caption{\textbf{OOD Detection on LIBERO Suite Shift (Goal $\rightarrow$ Spatial).}
We report planning-level AUROC for separating ID (Goal) and OOD (Spatial) using optimization-dynamics signals computed per planning call.
At a fixed operating point, \textbf{TPR} is specificity on ID instances and \textbf{TNR} is recall on OOD instances.
\textbf{Time Saved} is the average fraction of interaction time avoided on OOD episodes by early reporting, computed as $1 - t_{\mathrm{report}}/T_{\mathrm{total}}$.}
\label{tab:ood_comparison}
\vskip 0.1in
\begin{small}
\resizebox{\columnwidth}{!}{%
    \setlength{\tabcolsep}{4pt}
\begin{tabular}{l|c|cc|c}
\toprule
Method & AUROC ($\uparrow$) & TNR(\%)& TPR (\%)& \begin{tabular}{@{}c@{}}Time Saved(\%)\end{tabular} \\
\midrule
Baseline (Moving Avg) & 0.53 & 14.4 & 27.8  & 10.3 \\
Ours (Moving Avg) & \textbf{0.93} & 82.4  & 89.7  & 42.4 \\
\textbf{Ours (Leaky Bucket)} & \textbf{0.93} & \textbf{84.0} & \textbf{90.0} & \textbf{44.3} \\
\bottomrule
\end{tabular}%
}
\end{small}
\vskip -0.1in
\end{table}

\subsection{Ablations on Stationary Fields}
\label{subsec:ablation_alpha}

To evaluate the sensitivity of our adaptive optimization to the truncation schedule $c(\gamma)$, we ablate the parameter $\alpha$ (Table~\ref{tab:ablation_alpha}; see Appendix~\ref{sec:appendix_trunc} for further details). Crucially, removing truncation entirely ($\alpha = 1.0$) degrades the formulation into standard flow matching that merely discards the time conditioning. Without the velocity rescaling decay, the vector field fundamentally loses its equilibrium property: vector magnitudes no longer decay towards zero near the optimal state, leaving the optimization without a natural termination mechanism. This instability directly leads to the worst empirical performance across all tasks.

Conversely, the results highlight the strong hyperparameter robustness of our approach. While smaller values ($\alpha \le 0.6$) yield overly smooth fields that marginally slow convergence due to softer attractors, any choice of $\alpha < 1$ successfully preserves the essential equilibrium characteristics. As a result, all $\alpha < 1$ configurations maintain highly competitive success rates (averaging $>0.90$), provided the solver is allocated sufficient steps. Setting $\alpha = 0.8$ strikes the optimal balance between rapid convergence and precise policy control under a limited step budget.

\begin{table}[htb]
\centering
\caption{\textbf{Ablation Study on Truncation Parameter $\alpha$.} We report the performance on the LIBERO benchmark across different values of $\alpha$.}
\label{tab:ablation_alpha}
\vskip 0.1in
\begin{small}
\setlength{\tabcolsep}{4pt}
\begin{tabular}{lccccc}
\toprule
$\alpha$ & \textbf{Object} & \textbf{Goal} & \textbf{Spatial} & \textbf{Long} & \textbf{Avg} \\
\midrule
$\alpha = 0.0$ & \textbf{0.984} & 0.946 & 0.934 & 0.770 & 0.909 \\
$\alpha = 0.2$ & 0.968 & \textbf{0.954} & 0.958 & 0.802 & 0.921 \\
$\alpha = 0.4$ & 0.982 & 0.934 & 0.956 & 0.782 & 0.914 \\
$\alpha = 0.6$ & 0.982 & 0.942 & 0.956 & 0.778 & 0.915 \\
$\alpha = 0.8$ & 0.978 & 0.940 & \textbf{0.960} & \textbf{0.810} & \textbf{0.922} \\
$\alpha = 1.0$ & 0.932 & 0.840 & 0.898 & 0.558 & 0.807 \\
\bottomrule
\end{tabular}
\end{small}
\vskip -0.1in
\end{table}

\subsection{Real-World Experiment}
\label{sec:real_world}

To validate the practical effectiveness and efficiency of GeCO in unstructured physical environments, we conduct real-world benchmarks using the Galaxea R1 Lite\cite{jiang2025galaxea} mobile manipulator. 

\begin{figure*}[ht]
\centering
\includegraphics[width=2\columnwidth]{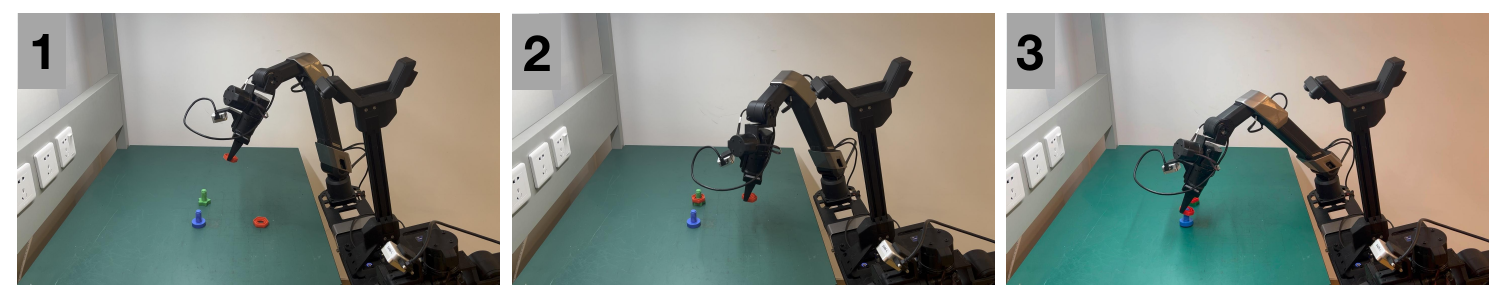} 
\caption{GeCO policy execution for the Nut Assembly task. The robot performs high-precision alignment and rotational insertion.}
\label{fig:nut_rollout}
\end{figure*}

\begin{figure*}[ht]
\centering
\includegraphics[width=2\columnwidth]{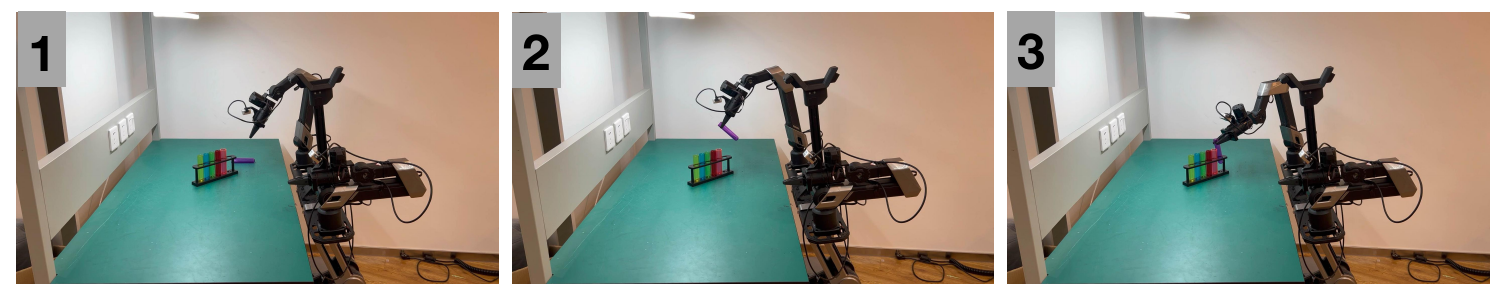} 
\caption{GeCO policy execution for the Chemistry Tube Arrangement task. The policy adaptively handles the tight-tolerance insertion of fragile tubes.}
\label{fig:tube_rollout}
\end{figure*}

\paragraph{Experimental Setup.} As shown in Fig.\ref{fig:nut_rollout} and Fig.\ref{fig:tube_rollout}, we use \textbf{Galaxea G0 Plus} \cite{jiang2025galaxea} as our baseline, which employs standard flow matching (FM) as VLA action head, and compare it against \textbf{G0 Plus with GeCO}. For both policies, the flow-matching heads are fine-tuned on a demonstration dataset collected via teleoperation. We select two complex and high-precision manipulation tasks for the benchmark: \textit{Nut Assembly} and \textit{Chemistry Tube Arrangement}. These tasks require millimeter-level precision, providing a rigorous testbed for GeCO's ability to perform deep refinement during contact-rich bottleneck phases while exiting early during simple transit motions. Each task is evaluated over 10 independent trials on the physical hardware. We control the same total gradient steps during training. Detailed experimental setup can be found in Appendix \ref{sec:appendix_real_world}. 

\paragraph{Quantitative Results.} 
As shown in Table~\ref{tab:real_world_results}, GeCO consistently outperforms the standard flow-matching baseline in success rate across both challenging real-world tasks, while also substantially reducing the average NFE and inference time. On \textit{Nut Assembly}, GeCO improves the success rate from 30\% to 70\%, reduces the average NFE from 10.0 to 5.06, and lowers the average inference time from 252.3 ms to 172.4 ms (a 31.7\% reduction). On \textit{Tube Arrangement}, GeCO improves the success rate from 25\% to 80\%, reduces the average NFE from 10.0 to 3.82, and lowers the average inference time from 254.6 ms to 150.3 ms (a 41.0\% reduction). While the baseline is constrained to a fixed schedule of 10 steps, GeCO's adaptive inference mechanism allows it to settle at equilibrium faster, thereby improving efficiency without compromising control fidelity (see Appendix~\ref{app:real_world_nfe} for detailed NFE analysis). We attribute this substantial performance margin to two key advantages: enhanced sample efficiency and superior generalization. First, given that both policies were fine-tuned using identical datasets and training budgets, GeCO's higher success rate clearly underscores the improved sample efficiency inherent to its time-unconditional formulation. Second, our method demonstrates remarkable robustness during physical deployment. Under domain-randomized evaluations, GeCO's optimization-based inference maintains robust convergence amidst real-world visual and physical variations, whereas the fixed-schedule baseline proves brittle to such unstructured domain shifts.

\begin{table*}[!t]
\caption{Real-world performance on Galaxea R1 Lite. We report the success rate, average Number of Function Evaluations (NFE), and average inference time for each task.}
\label{tab:real_world_results}
\begin{small}
\centering
\resizebox{\textwidth}{!}{%
    \setlength{\tabcolsep}{4pt}
    \begin{tabular}{l|ccc|ccc}
    \toprule
    Method 
    & \multicolumn{3}{c|}{Nut Assembly} 
    & \multicolumn{3}{c}{Tube Arrangement} \\
    & Success ($\uparrow$) & Avg. NFE ($\downarrow$) & Avg. Infer. Time ($\downarrow$)
    & Success ($\uparrow$) & Avg. NFE ($\downarrow$) & Avg. Infer. Time ($\downarrow$) \\
    \midrule
    G0 Plus 
    & 30\% & 10.0 & 252.3 
    & 25\% & 10.0 & 254.6 \\
    \textbf{G0 Plus + GeCO (Ours)} 
    & \textbf{70\%} & \textbf{5.06} & \textbf{172.4}
    & \textbf{80\%} & \textbf{3.82} & \textbf{150.3} \\
    \bottomrule
    \end{tabular}%
}
\end{small}
\vskip -0.1in
\end{table*}

\section{Discussion and Limitations}\label{sec:discussion}
GeCO reformulates generative action synthesis as a robust optimization task, and the proposed framework delivers promising performance alongside strong adaptivity and OOD awareness, which validates the effectiveness of our core design. While the framework performs well in practical settings, we acknowledge that its theoretical foundation can be further strengthened. Future work will focus on deriving more rigorous theoretical bounds for performance gains, the convergence speed of the optimization process, and the OOD detection mechanism in the context of control and VLA to further solidify GeCO’s theoretical foundation.
\section{Conclusion}\label{sec:conclusion}
We introduced \textbf{GeCO}, a time-unconditional formulation of flow-matching policies that casts action generation as iterative optimization over a stationary velocity field, where expert behaviors form stable attractors via velocity rescaling. 
This removes the rigid dependence on fixed integration schedules and provides a plug-and-play replacement for continuous action heads in modern VLA systems. 
Experimentally, GeCO delivers a better efficiency--performance trade-off through state-dependent adaptive optimization, scales to flow matching based VLA models, and yields an optimization-native safety signal: the final field norm serves as a training-free OOD score. 
\section*{Impact Statement}

This paper aims to advance the efficiency and safety of generative robotic control. The proposed framework, GeCO, introduces an intrinsic anomaly detection mechanism that helps mitigate operational risks in unstructured environments. Additionally, our adaptive inference approach optimizes computational resource usage, enabling more efficient deployment on robot hardware. We do not foresee specific negative societal consequences beyond the general implications of advancing automation technology.

\bibliography{example_paper}
\bibliographystyle{icml2026}

\newpage
\appendix
\onecolumn



\section{Implementation Details for Adaptive Inference and Training}
\label{sec:appendix_impl_adaptive}

\subsection{Codebase and Baselines}
\label{sec:appendix_codebase}
All experiments are implemented on top of the CleanDiffuser \cite{dong2024cleandiffuser} codebase.
We use the same network backbone as a Continuous Rectified Flow policy \cite{liu2022flow}, but differ in the learning objective and the test-time inference algorithm. 

\subsection{Model Architecture}
\label{sec:appendix_arch}
We follow the Rectified Flow architecture with a 1D diffusion-style transformer and a frozen vision-language conditioner.

\paragraph{Vision-language condition.}
We use a frozen DINOv2 \cite{oquab2023dinov2} vision encoder and a T5 \cite{raffel2020exploring} text encoder. The T5 hidden dimension is set by the pretrained configuration:
\[
\texttt{t5\_hidden\_dim} \leftarrow \texttt{T5Config.from\_pretrained(t5\_model).d\_model}.
\]
The conditioner is instantiated as:
\begin{itemize}
    \item \texttt{ViTAndT5VisionLanguageCondition} with \texttt{emb\_dim=768}, \texttt{freeze=True}, \texttt{To=1}, and \texttt{n\_views=2}.
\end{itemize}

\paragraph{Policy backbone.}
The action generator uses a DiT-style 1D transformer with cross-attention conditioning:
\begin{itemize}
    \item \texttt{DiT1dWithACICrossAttention} with \texttt{x\_dim=act\_dim}, \texttt{x\_seq\_len=Ta(=16)}, \texttt{emb\_dim=768}, \texttt{d\_model=384},
    \texttt{n\_heads=6}, \texttt{depth=12}.
\end{itemize}

GeCO uses the same backbone modules for fair comparison, while changing the training objective and inference procedure.

\subsection{Adaptive Inference Setup}
\label{sec:appendix_adaptive_infer}

\paragraph{Sampler.}
We use a standard gradient-descent (GD) sampler with a maximum of $K_{\max}=30$ refinement steps. Denoting the refinement direction by $f_\theta(a_k, s)$, the update is:
\[
a_{k+1} \leftarrow a_k - \eta_k \, f_\theta(a_k, s), \quad k=0,\dots,K_{\max}-1.
\]

\paragraph{Step-size schedule.}
We use a fixed per-step learning-rate schedule $\{\eta_k\}_{k=1}^{30}$:
\[
\eta_{1}=0.1,\quad
\eta_{2..4}=0.05,\quad
\eta_{5..16}=0.02,\quad
\eta_{17..30}=0.01.
\]
We enable adaptive early stopping using a norm-based stopper: we terminate refinement once
\[
\|f_\theta(a_k,s)\|_2 < 0.4.
\]

\subsection{Training Recipe and Hyperparameters}
\label{sec:appendix_train_recipe}

\paragraph{Pretrained encoders.}
\begin{itemize}
    \item T5: \texttt{google-t5/t5-base}
    \item ViT: \texttt{facebook/dinov2-base}
\end{itemize}

\paragraph{Training details.}
We use a truncated linear decay for equilibrium field learning
\label{sec:appendix_trunc}
\begin{equation}
c(\gamma)
=
\begin{cases}
\lambda, & \gamma \le \alpha_0, \\
\lambda \dfrac{1-\gamma}{1-\alpha_0}, & \gamma > \alpha_0,
\end{cases}
\end{equation}
with scale $\lambda = 4$ and onset $\alpha_0 = 0.1$.

\paragraph{Constants.}
\begin{itemize}
    \item Observation horizon: $T_o=1$
    \item Action horizon: $T_a=16$
    \item Executed actions per step: \texttt{num\_act\_exec=8}
    \item Normalization parameters: \texttt{norm\_params=[0.5, 0.5, 0.5]}
\end{itemize}

\section{Benchmarks}\label{sec:appendix_benchmarks}
\paragraph{LIBERO.}
We evaluate on the LIBERO simulation benchmark~\cite{liu2023libero}, which contains four task suites---\textit{Goal}, \textit{Spatial}, \textit{Object}, and \textit{Long}---designed to isolate different distribution shifts in language-conditioned manipulation.
\textit{Goal} varies task goals while keeping objects and layouts fixed; \textit{Spatial} introduces unseen spatial configurations; \textit{Object} shifts object categories; and \textit{Long} consists of long-horizon, compositional tasks.
For policy learning and evaluation, our model predicts action chunks of length $T_a=16$; after each prediction, we execute the first $8$ actions (\texttt{num\_act\_exec}=8) in closed loop before replanning the next chunk.
The observation space consists of two-view RGB images at the current time step with $T_o=1$ (no history), and the language instruction is encoded by a frozen T5 encoder (\texttt{google-t5/t5-base}).
Unless otherwise specified, we train each suite-specific policy for 30k optimization steps and report success rates averaged over 50 rollouts, following the standard evaluation protocol of LIBERO.

\paragraph{RoboTwin 2.0.}
We additionally benchmark on RoboTwin 2.0~\cite{chen2025robotwin}, a bimanual manipulation benchmark that evaluates robustness under controlled domain shifts.
RoboTwin 2.0 reports performance under two difficulty settings: \textit{Easy} uses clean environments, while \textit{Hard} introduces domain-randomized perturbations such as clutter distractors, texture changes, lighting variations, and tabletop-height shifts.
We evaluate on five representative tasks (\textit{Adjust Bottle}, \textit{Beat Block Hammer}, \textit{Blocks Ranking (Size)}, \textit{Click Alarmclock}, and \textit{Click Bell}) and report Easy/Hard success and their averages (Table~\ref{tab:robotwin_benchmark_vla_eh}).
All methods are evaluated with 100 rollouts per setting, using the official success criteria of RoboTwin 2.0.
We highlight that the Hard setting creates a realistic robustness test by perturbing both visual appearance and scene dynamics while keeping task semantics unchanged.

\paragraph{VLABench.}
We further evaluate on VLABench~\cite{zhang2025vlabench}, a large-scale benchmark for language-conditioned manipulation that emphasizes long-horizon reasoning and semantic grounding.
VLABench organizes tasks into multiple tracks covering in-distribution execution, cross-category generalization, common-sense/world-knowledge transfer, semantic instruction understanding, and unseen texture generalization.
We report track-wise success rates on Tracks and their average (Table~\ref{tab:vlabench_benchmark_vla}), following the benchmark’s standard evaluation protocol with 50 rollouts. We include per-task breakdown for the 10-task primitive subset in Table~\ref{tab:vlabench_per_task_full}, which complements the track-level summary in Table~\ref{tab:vlabench_benchmark_vla}.

\begin{table*}[t]
\centering
\caption{Per-task success rates on VLABench primitive 10-task evaluation. We report success rates for the baseline $\pi_0$ and $\pi_0$ with GeCO (ours) across five tracks; higher is better. Track averages correspond to the averages over the 10 tasks in each track.}
\label{tab:vlabench_per_task_full}
\vspace{-2pt}
\begin{small}
\setlength{\tabcolsep}{3.5pt}
\renewcommand{\arraystretch}{1.06}
\begin{tabular}{l|cc|cc|cc|cc|cc}
\toprule
\multirow{2}{*}{Task}
& \multicolumn{2}{c|}{Track 1 }
& \multicolumn{2}{c|}{Track 2 }
& \multicolumn{2}{c|}{Track 3}
& \multicolumn{2}{c|}{Track 4 }
& \multicolumn{2}{c}{Track 6 } \\
\cmidrule(lr){2-3}\cmidrule(lr){4-5}\cmidrule(lr){6-7}\cmidrule(lr){8-9}\cmidrule(lr){10-11}
& $\pi_0$ & Ours
& $\pi_0$ & Ours
& $\pi_0$ & Ours
& $\pi_0$ & Ours
& $\pi_0$ & Ours \\
\midrule
add\_condiment         & 0.66  & 0.80  & 0.14  & 0.08  & 0.34  & 0.16  & 0.26  & 0.10  & 0.56  & 0.76 \\
insert\_flower         & 0.18  & 0.30  & 0.04  & 0.00  & 0.22  & 0.28  & 0.02  & 0.00  & 0.10  & 0.02 \\
select\_book           & 0.694 & 0.84  & 0.064 & 0.08  & 0.417 & 0.51  & 0.311 & 0.23  & 0.714 & 0.79 \\
select\_chemistry\_tube& 0.52  & 0.76  & 0.12  & 0.14  & 0.70  & 0.94  & 0.06  & 0.02  & 0.28  & 0.71 \\
select\_drink          & 0.52  & 0.78  & 0.224 & 0.30  & 0.08  & 0.10  & 0.10  & 0.04  & 0.44  & 0.66 \\
select\_fruit          & 0.38  & 0.56  & 0.46  & 0.52  & 0.083 & 0.14  & 0.06  & 0.18  & 0.30  & 0.32 \\
select\_mahjong        & 0.25  & 0.58  & 0.02  & 0.04  & 0.125 & 0.12  & 0.12  & 0.06  & 0.02  & 0.25 \\
select\_painting       & 0.46  & 0.22  & 0.26  & 0.12  & 0.50  & 0.60  & 0.56  & 0.76  & 0.30  & 0.18 \\
select\_poker          & 0.54  & 0.70  & 0.26  & 0.54  & 0.06  & 0.22  & 0.12  & 0.52  & 0.28  & 0.52 \\
select\_toy            & 0.50  & 0.52  & 0.36  & 0.34  & 0.38  & 0.32  & 0.12  & 0.10  & 0.18  & 0.28 \\
\midrule
Avg\_SR                & 0.47  & 0.61  & 0.212 & 0.22  & 0.291 & 0.34  & 0.173 & 0.20  & 0.322 & 0.45 \\
\bottomrule
\end{tabular}
\end{small}
\vspace{-6pt}
\end{table*}

\section{VLA Training Details.}
We fine-tune $\pi_0$ / $\pi_{0.5}$ VLAs using the OpenPI training stack with LeRobot-format datasets. Across all benchmarks, we train for \textbf{30k} gradient steps . Unless otherwise specified, weights are initialized from the corresponding OpenPI base checkpoint.

\paragraph{RoboTwin 2.0.}
For RoboTwin 2.0, we fine-tune a separate $\pi_0$ model per task. Observations include three RGB views (\texttt{cam\_high}, \texttt{cam\_left\_wrist}, \texttt{cam\_right\_wrist}) and robot proprioception (\texttt{observation.state}).
We train for 30k steps with batch size 32 for each task.

\paragraph{LIBERO.}
For $\pi_0$ on LIBERO, we fine-tune on \texttt{physical-intelligence/libero} datasets. We initialize from \texttt{gs://openpi-assets/checkpoints/pi0\_base/params} and train for 30k steps with batch size 32.

For $\pi_{0.5}$ on LIBERO, we fine-tune for 30k steps with batch size 256. We use AdamW with gradient clipping (\texttt{clip\_gradient\_norm=1.0}) and an EMA of model weights (\texttt{ema\_decay=0.999}). The learning rate follows a cosine schedule with warmup (\texttt{warmup\_steps=10{,}000}, \texttt{peak\_lr=5e-5}). 

\paragraph{VLABench.}
For the VLABench fine-tuning experiment, we choose to fine-tune $\pi_0$ on huggingface datasets \texttt{vlabench/vlabench\_primitive\_ft\_lerobot}. We trained for 30k steps with batch size 256.
\section{OOD Detection Settings}
\label{sec:appendix_ood}

\paragraph{ID/OOD split.}
We study a controlled suite-level distribution shift on LIBERO by treating \textit{LIBERO-Goal} as in-distribution (ID) and \textit{LIBERO-Spatial} as out-of-distribution (OOD). We train a single policy on the 10 tasks of LIBERO-Goal and evaluate OOD detection on the 10 tasks of LIBERO-Spatial.

\paragraph{Episode and planning-call sampling.}
We sample 500 ID episodes from LIBERO-Goal and 500 OOD episodes from LIBERO-Spatial. Each episode consists of multiple planning calls, where the policy generates an action chunk and executes a fixed number of actions before replanning.
To avoid overweighting longer episodes, we compute planning-level statistics using a fixed number of planning calls per episode: we extract $J=\texttt{20}$ planning calls from each episode using random $J$ calls and treat each extracted planning call as one evaluation instance.

\paragraph{GeCO score (optimization-dynamics signal).}
For each planning call at state $s^{(j)}$, GeCO performs iterative optimization for at most $K_{\max}=10$ steps and produces per-step norms $r_k=\|f_\theta(a_k,s^{(j)})\|_2$. We use the final-step norm as the raw anomaly score for that planning call:
\[
\text{score}(\text{plan } j)=\|f_\theta(a^{(j)}_{K}, s^{(j)})\|_2,
\]
where $K\le K_{\max}$ is the last executed refinement step (early termination uses the stopping step; otherwise $K=K_{\max}$).

\paragraph{Temporal filters and triggering rule.}
We evaluate two online filters over $\{r_k\}_{k=1}^{K}$ within each planning call:
(i) \textbf{Moving Average} with window $w=5$, triggering when $\tilde r_k>\tau_{\mathrm{ma}}$;
(ii) \textbf{Leaky Bucket} defined by
$
b_k \leftarrow \max\{0,(1-\lambda)b_{k-1}+(r_k-\tau)_+\},
$
triggering when $b_k\ge B$.
We set $(\tau_{\mathrm{ma}},\tau,\lambda,B)=\texttt{(0.6, 0.5, 0, 0.7)}$ and keep them fixed across all experiments.

\paragraph{Baseline score (training-loss proxy).}
As a baseline OOD signal, we follow the Diff-DAgger-style idea~\cite{lee2025diff} of using the policy’s training objective as an uncertainty indicator. For each planning call, let $x_0$ denote the generated action chunk and let $c$ denote the conditioning input. We draw a single Gaussian noise sample $\epsilon \sim \mathcal{N}(0,I)$ and construct a noised input at timestep $t$ via linear interpolation
\[
x_t \;=\; (1-t)\,x_0 + t\,\epsilon,
\]
where $T$ is the total number of diffusion steps used by the training objective.
We then evaluate the flow-matching regression loss
\[
\mathcal{L}_{\mathrm{FM}}(x_0, c)
=
\mathbb{E}\left\| f_\theta(x_t, t, c) - (\epsilon - x_0) \right\|_2^2
\]

\paragraph{AUROC computation.}
We compute planning-level AUROC by sweeping a threshold over the scalar score to separate OOD planning calls (positive) from ID planning calls (negative), using the $500\times J$ ID instances and $500\times J$ OOD instances.

\paragraph{Operating point (TNR/TPR).}
We report TNR/TPR at a single operating point selected to achieve \textbf{TPR $\approx 90\%$} on ID instances for our detector when possible. Specifically, we choose the threshold $\theta$ such that $\mathrm{TPR}(\theta)$ is closest to 0.9 and report the corresponding TNR.

\paragraph{Early reporting and Time Saved.}
The detector runs online during an episode. Once any planning call triggers an OOD flag, we immediately terminate the episode.
We define the reporting time $t_{\mathrm{report}}$ as the environment time step when the first trigger occurs, and compute
\[
\text{Time Saved} = 1-\frac{t_{\mathrm{report}}}{T_{\mathrm{total}}},
\]
where $T_{\mathrm{total}}=300$ is the episode horizon under the standard evaluation protocol. We report Time Saved averaged over OOD episodes only.

\section{Real World Experiment Settings}
\label{sec:appendix_real_world}

\paragraph{Hardware Configuration.}
We conduct our real-world experiments using the \textbf{Galaxea R1 Lite} mobile manipulator. The robot is equipped with two dexterous arms and 3 RGB cameras. For our policy input, we utilize three camera views: a head camera (\texttt{cam\_head}), and two wrist-mounted cameras (\texttt{cam\_left\_wrist}, \texttt{cam\_right\_wrist}). We also include the robot's proprioceptive state (\texttt{observation.state}), which contains joint positions and gripper states. The control frequency of the physical system is set to 15 Hz, and the execution action chunk size is set to $T_a = 16$.

\begin{figure*}[h]
  \centering
  \includegraphics[width=0.8\textwidth]{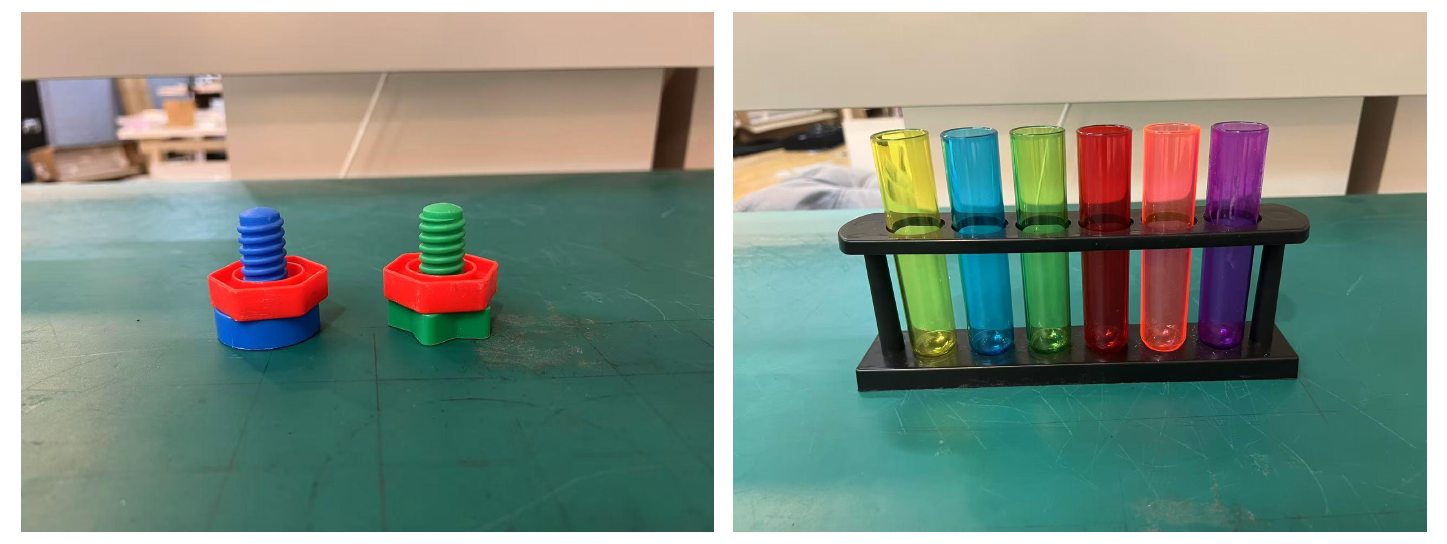}
  \caption{Task setups for the real-world robotic deployment, showing the configurations for both the Nut Assembly and the Chemistry Tube Arrangement tasks.}
  \label{fig:experimental_setups}
\end{figure*}

\paragraph{Data Collection and Task Description.}
We select two contact-rich, high-precision manipulation tasks: \textit{Nut Assembly} and \textit{Chemistry Tube Arrangement}. The task setup is shown in Fig. \ref{fig:experimental_setups}.

\begin{itemize}
    \item \textbf{Nut Assembly:} The robot shall grasp a plastic nut, accurately align it with a threaded bolt, and perform the insertion. A trial is considered successful only when both two nuts are fully threaded onto their respective bolts.
    \item \textbf{Chemistry Tube Arrangement:} The robot shall grasp chemistry tubes from a randomized initial pose and insert them into designated tight-tolerance slots on a rack with random position and rotation. A successful trial requires the tube to be fully seated in the correct slot without dropping or tipping over.
\end{itemize}

\paragraph{Training Details.}
Our baseline policy is \textbf{Galaxea G0 Plus}, which employs a standard flow-matching-based continuous action head. To train our \textbf{G0 Plus with GeCO} policy, we initialize the model with the pre-trained G0 Plus weights. Following the GeCO formulation, we fix the time conditioning to $t=0$ during fine-tuning to learn a stationary velocity field. We fine-tune the model on the expert demonstrations for 4 full epochs with a batch size of 32. We use the AdamW optimizer with a peak learning rate of \texttt{2.5e-5} and a cosine learning rate schedule with \texttt{warmup\_steps=200}. 

\paragraph{GeCO Inference Setup.}
During real-world deployment, the GeCO policy synthesizes action chunks via iterative optimization. We set the maximum number of optimization steps to $K_{\max} = 10$. We use a step-size schedule $\{\eta_k\}$ initialized at \texttt{0.25} and decaying to \texttt{0.02} for deeper refinement. The adaptive early stopping mechanism is triggered when the gradient norm falls below a threshold $\tau_{\mathrm{opt}}$:
\[
\|f_\theta(a_k, s_t)\|_2 < \tau_{\mathrm{opt}},
\]
where $\tau_{\mathrm{opt}}$ is empirically set to \texttt{1.5}. If the threshold is not met, the optimization continues until $K_{\max}$ is reached, ensuring complex states (e.g., insertion alignment) receive maximum computational budget while simple transits exit early.

\paragraph{Evaluation Protocol.}
We evaluate both the baseline (G0 Plus) and our method (G0 Plus + GeCO) over 20 independent physical trials for each task. The initial positions of the objects (nuts, bolts, tubes, racks) are randomized within the workspace to test robustness to spatial variations. We report the success rate and record the average Number of Function Evaluations (NFE) per planning call. For the baseline, inference is bound to a fixed schedule of NFE $= 10$. For GeCO, NFE varies dynamically based on the early stopping criteria.

\section{Detailed Analysis of Real-World Adaptive Computation}
\label{app:real_world_nfe}

\begin{figure}[htbp]
    \centering
    \includegraphics[width=\columnwidth]{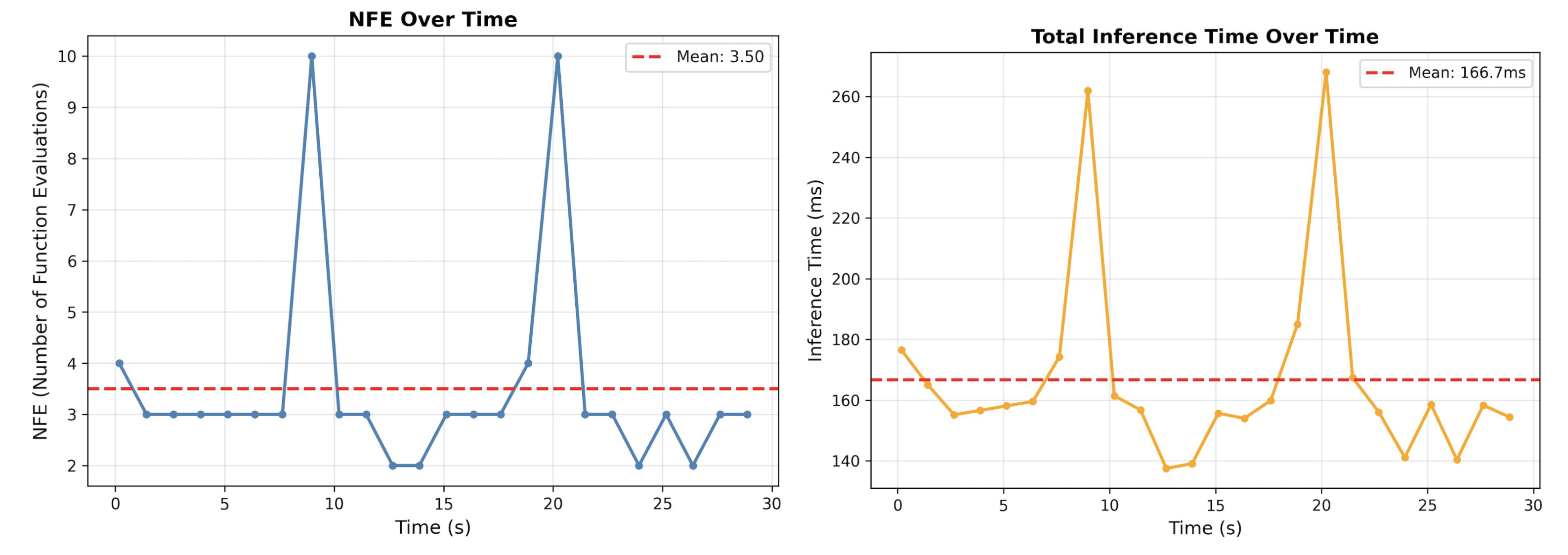}
    \includegraphics[width=\columnwidth]{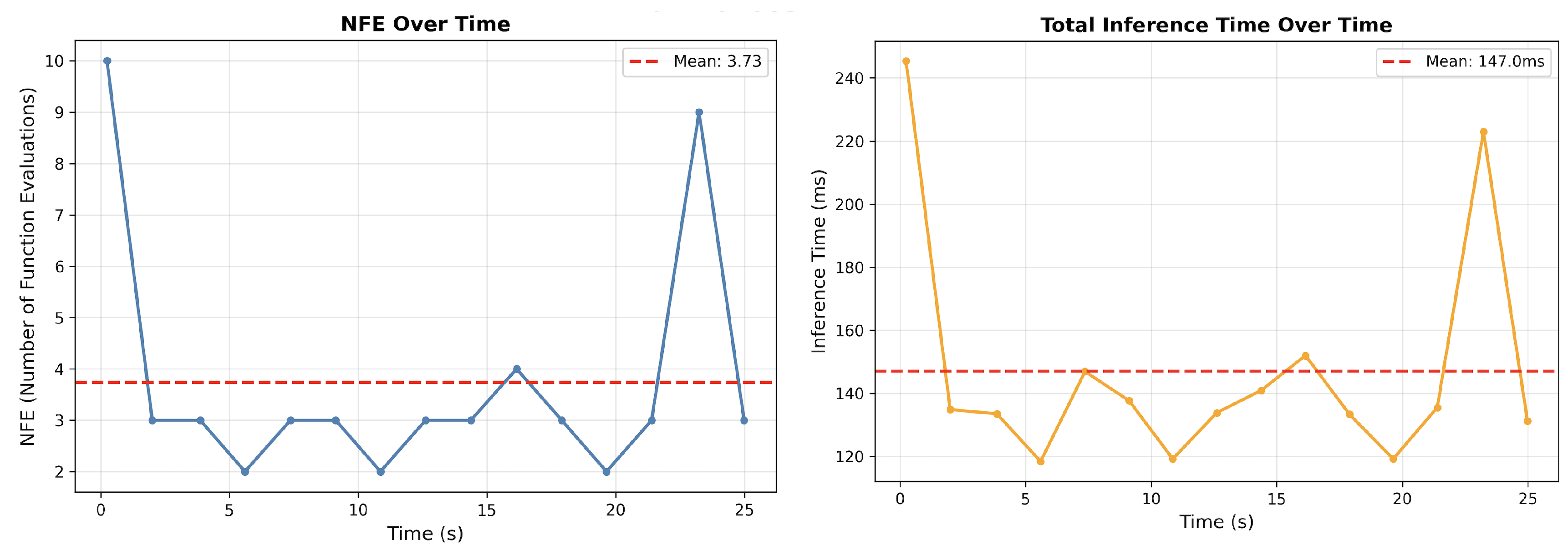}
    \caption{\textbf{Adaptive computation during real-world execution.} We visualize the per-step Number of Function Evaluations (NFE) and total inference time over time for representative physical rollouts of \textit{Nut Assembly} and \textit{Tube Arrangement}. For \textit{Nut Assembly}, GeCO uses an average of 3.50 NFEs and 166.7\,ms total inference time per decision. For \textit{Tube Arrangement}, GeCO uses an average of 3.73 NFEs and 147.0\,ms per decision. Rather than relying on a fixed computation budget, GeCO automatically increases computation during bottleneck phases, where the NFE rises to 9--10, and exits early during easier free-space motions, where the NFE drops to as low as 2--3. The corresponding inference-time traces mirror this behavior, showing that GeCO adaptively allocates computation according to task difficulty and thereby improves real-world inference efficiency.}
    \label{fig:appendix_real_world_nfe}
\end{figure}

To provide a deeper understanding of how GeCO dynamically modulates its computation during physical deployment, we present a detailed analysis of both the Number of Function Evaluations (NFE) and the total inference time over representative real-world rollouts in Fig.~\ref{fig:appendix_real_world_nfe}.

As shown in Fig.~\ref{fig:appendix_real_world_nfe}, peaks in computation coincide with the most challenging stages of execution, such as fine alignment, grasping, and contact-rich insertion. In contrast, during simpler semantic stages and free-space transit, the policy quickly converges and requires substantially fewer function evaluations. The total inference time follows the same overall trend as the NFE profile, confirming that GeCO's adaptive stopping behavior directly translates into lower online compute cost on physical hardware. These results further demonstrate that our time-unconditional flow matching formulation enables state-dependent computation in real-world manipulation while maintaining robust performance.

\section{Additional Ablation Studies}
\label{app:ablation}

In this section, we provide additional ablation studies to further validate the design choices and hyperparameters of our proposed method.

\subsection{Ablation Study: Effect of Truncation Parameter $\alpha$}
\label{subsec:ablation_alpha_vla} 

While Section~\ref{subsec:ablation_alpha} validates the stationary field formulation primarily using the DiT architecture, it is crucial to verify that our design choices generalize across different network backbones. Therefore, in this section, we extend our analysis by conducting an ablation study on the truncation parameter $\alpha$ specifically using the VLA architecture. As discussed in Section~\ref{subsec:uncond-field}, $\alpha$ controls the onset of the velocity rescaling decay, directly impacting the equilibrium convergence. Table~\ref{tab:ablation_alpha} presents the success rates of the VLA-based policy on the LIBERO benchmark across different values of $\alpha \in \{0.6, 0.8, 0.9\}$. 

The results highlight a clear trade-off governed by the smoothness of the learned vector field. When $\alpha$ is too small (e.g., $\alpha=0.6$), the field becomes overly smooth. While this provides stable gradient directions, it drastically slows down convergence. The optimization often fails to reach the tight equilibrium manifold within a limited step budget. Conversely, an excessively large $\alpha$ (e.g., $\alpha=0.9$) leads to a large Lipschitz constant ($L$-smoothness) for the velocity field. This creates a sharp and stiff optimization landscape, making it difficult for the iterative solver to converge stably, thus degrading overall performance. Setting $\alpha=0.8$ strikes the optimal balance, ensuring the attractors are sharp enough for precise control while maintaining a smooth enough field for rapid and stable inference.

\begin{table}[htb]
\centering
\caption{\textbf{Ablation Study on Truncation Parameter $\alpha$ (VLA Architecture).} We report the success rates (\%) on the LIBERO benchmark across different values of $\alpha$ to demonstrate its impact on policy performance.}
\label{tab:ablation_alpha_vla}
\vskip 0.1in
\begin{small}
\setlength{\tabcolsep}{4pt}
\begin{tabular}{l|cccc|c}
\toprule
\multirow{2}{*}{$\alpha$} & \multicolumn{5}{c}{Success Rate (\%)} \\
\cmidrule{2-6}
 & Goal & Spat. & Obj. & Long & Avg. \\
\midrule
$0.6$ & 85.8 & \textbf{98.0} & 91.6 & 48.6 & 81.0 \\
$0.8$ & \textbf{96.4} & 97.4 & \textbf{98.6} & \textbf{92.4} & \textbf{96.2} \\
$0.9$ & 95.8 & 97.8 & 94.0 & 81.2 & 92.2 \\
\bottomrule
\end{tabular}
\end{small}
\vskip -0.1in
\end{table}

\subsection{Ablation Study: Effect of the Early-Stop Threshold $\tau$}
\label{subsec:ablation_tau}

To investigate the trade-off between policy performance and inference speed, we ablate the early-stop threshold $\tau$. Table~\ref{tab:ablation_tau} summarizes the success rates and the average number of inference steps required. A smaller threshold (e.g., $\tau = 0.4$) enforces stricter convergence criteria, allowing the solver to approach the exact equilibrium. This yields the highest average success rate ($93.5\%$) but requires more inference steps ($11.6$ steps on average). Conversely, relaxing the threshold to $\tau = 0.8$ significantly accelerates inference, reducing the average steps to $6.6$, with only a marginal drop in overall performance ($91.9\%$ average success). This demonstrates that our early-stopping mechanism provides a flexible and robust knob to balance task accuracy and computational cost during real-world deployment.

\begin{table}[htb]
\centering
\caption{\textbf{Ablation on the Early-Stop Threshold $\tau$.} We report the success rates (\%) and the average number of inference steps across the LIBERO benchmark.}
\label{tab:ablation_tau}
\vskip 0.1in
\begin{small}
\setlength{\tabcolsep}{4pt}
\begin{tabular}{lcccccc}
\toprule
$\tau$ & \textbf{LIBERO Object} & \textbf{LIBERO Goal} & \textbf{LIBERO Spatial} & \textbf{LIBERO Long} & \textbf{Avg} & \textbf{Avg. Steps} \\
\midrule
$0.4$ & 99.0 & 95.2 & 95.8 & 83.8 & 93.5 & 11.6 \\
$0.6$ & 98.8 & 94.0 & 95.6 & 81.0 & 92.35 & 7.7 \\
$0.8$ & 96.8 & 93.2 & 96.0 & 81.6 & 91.9 & 6.6 \\
\bottomrule
\end{tabular}
\end{small}
\vskip -0.1in
\end{table}

\subsection{Ablation Study: Robustness to Fixed Step-Size Schedules}
\label{subsec:ablation_step_size}

Finally, we ablate the step-size schedule to evaluate its impact on optimization stability. Table~\ref{tab:ablation_step} compares our current schedule against various fixed step-size baselines (e.g., applying a constant step size for 20 steps). The empirical results show that even simple fixed step-size schedules remain highly competitive. For instance, the $0.02 \times 20$ schedule achieves an impressive $93.05\%$ average success rate, which is only marginally lower than our current schedule ($93.5\%$). The fact that our current schedule provides only a modest improvement over several reasonable alternatives is a crucial finding. It strongly suggests that GeCO's performance gains stem primarily from the fundamental stationary-field formulation itself, rather than relying on a highly specialized or fragile learning-rate schedule.

\begin{table}[htb]
\centering
\caption{\textbf{Ablation on Fixed Step-Size Schedules.} We compare our current adaptive schedule with several fixed step-size schedules (constant step size $\times$ total steps) on the LIBERO benchmark.}
\label{tab:ablation_step}
\vskip 0.1in
\begin{small}
\setlength{\tabcolsep}{4pt}
\begin{tabular}{lccccc}
\toprule
\textbf{Step-size schedule} & \textbf{LIBERO Object} & \textbf{LIBERO Goal} & \textbf{LIBERO Spatial} & \textbf{LIBERO Long} & \textbf{Avg} \\
\midrule
$0.015 \times 20$ & 97.0 & 92.0 & 94.8 & 77.2 & 90.3 \\
$0.02 \times 20$ & 98.0 & 94.8 & 95.4 & 84.0 & 93.1 \\
$0.025 \times 20$ & 98.4 & 95.0 & 94.2 & 80.4 & 92.0 \\
$0.035 \times 20$ & 98.0 & 95.0 & 95.0 & 77.4 & 91.4 \\
\midrule
Current schedule & 99.0 & 95.2 & 95.8 & 83.8 & 93.5 \\
\bottomrule
\end{tabular}
\end{small}
\vskip -0.1in
\end{table}

\end{document}